\documentclass{article}

\usepackage{soul}
\usepackage{PRIMEarxiv}
\usepackage{authblk}
\usepackage[utf8]{inputenc} 
\usepackage[T1]{fontenc}    
\usepackage{hyperref}       
\usepackage[superscript,sort,compress,nomove]{cite}
\usepackage{url}            
\usepackage{booktabs}       
\usepackage{multirow}
\usepackage{graphicx}
\usepackage{amssymb}
\usepackage{amsfonts}       
\usepackage{xcolor}         
\usepackage{nicefrac}       
\usepackage{microtype}      
\usepackage{lipsum}
\usepackage{fancyhdr}       
\usepackage{graphicx}       
\graphicspath{{media/}}     
\usepackage{amsmath}
\usepackage{dsfont}
\usepackage{comment}

\usepackage{booktabs}
\usepackage{tabularx}
\usepackage{array} 
\usepackage{subcaption}

\usepackage{makecell}    
\usepackage{adjustbox}   
\usepackage{float}       

\title{Single vs. Multiple Branches in DeepONet and S-DeepONet: Network Architecture Follows Coupling in Multiphysics Systems}

\author[1,2]{Jaewan Park$^{\dagger}$}
\author[3]{Kazuma Kobayashi$^{\dagger}$}
\author[1,4]{Qibang Liu}
\author[1,5]{Diab Abueidda}
\author[1,2]{Seid Koric}
\author[1,3]{Syed Bahauddin Alam\textsuperscript{*}}

\affil[1]{National Center for Supercomputing Applications, University of Illinois at Urbana-Champaign, 1205 W. Clark St., Urbana, IL, 61801, USA
}

\affil[2]{The Grainger College of Engineering, Mechanical Science and Engineering, University of Illinois at Urbana-Champaign, 1206 W Green St, Urbana, IL 61801, USA
}

\affil[3]{The Grainger College of Engineering, Nuclear, Plasma \& Radiological Engineering, University of Illinois at Urbana-Champaign, 104 South Wright Street, Urbana, IL 61801, USA
}

\affil[4]{Department of Industrial and Manufacturing Systems Engineering, Kansas State University, Manhattan, KS 66506, USA}

\affil[5]{Civil and Urban Engineering Department, New York University Abu Dhabi, UAE
}

\affil[$\dagger$]{These authors contributed equally to this work}

\affil[*]{Corresponding author: \href{mailto:alams@illinois.edu}{alams@illinois.edu}}

\begin{document}
\maketitle

\begin{abstract}
Real-time prediction of complex physical systems requires surrogate models that learn from data while representing strong multiphysics coupling. Deep Operator Networks have shown success in single-physics problems, yet their effectiveness in capturing nonlinear interactions in coupled systems (such as thermo-mechanical or electro-thermal coupling) remains underexplored. Here we pose a practical question: \textit{should the architecture of a neural operator reflect the strength of physical coupling it aims to model?} We compare single-branch and multi-branch designs, in both feedforward and sequential recurrent forms, across three representative systems: a reaction–diffusion problem with heterogeneous sources, a nonlinear thermo-electrical problem with temperature-dependent conductivity and Joule heating, and a viscoplastic thermo-mechanical model of steel solidification. Single-branch networks consistently outperform multi-branch variants in tightly coupled regimes by encouraging shared latent representations, whereas multi-branch designs remain favorable for decoupled or single-physics tasks. Once trained, these surrogates deliver full-field predictions up to $\mathbf{1.8 \times 10^4}$ times faster than physics-based solvers.

\end{abstract}


\section*{Introduction}
Machine learning (ML) has rapidly transformed computational science by enabling efficient data-driven approximations of complex physical phenomena. Artificial neural networks (NNs), inspired by biological neural systems, have been particularly successful in physics-based modeling, where they serve as surrogates for computationally expensive numerical solvers. Trained on representative datasets, these models approximate input–output mappings of governing equations, supporting tasks such as sensitivity analysis, uncertainty quantification \cite{kobayashi2024ai,kumar2022multi}, and inverse design across domains including cardiac electrophysiology \cite{pagani2021enabling}, optics \cite{shahane2022surrogate}, and advanced manufacturing \cite{liu2024adaptive, kushwaha2024advanced, goli2020chemnet}, multi-physics engineering \cite{kollmann2020deep, bastek2023inverse}.

To reduce reliance on large training datasets, physics-informed neural networks (PINNs) integrate governing physical laws directly into their loss functions \cite{raissi2019physics, sun2020surrogate, he2023deep, fuhg2022mixed, nguyen2020deep}. \textcolor{black}{By embedding partial differential equations (PDEs) and boundary conditions, PINNs can train with limited labeled data. Despite this promise, PINNs often suffer from poor scalability in high-dimensional settings, high training cost, and reduced accuracy in strongly coupled or path-dependent problems \cite{krishnapriyan2021characterizing, wang2022and}. These limitations have motivated exploration of alternative frameworks that retain physical fidelity while improving scalability.}

Neural operator learning has emerged as such an alternative. Rather than mapping finite-dimensional vectors, neural operators approximate nonlinear mappings between infinite-dimensional function spaces \cite{kovachki2023neural}. This operator perspective enables predictions of entire solution fields across arbitrary discretizations, offering geometric flexibility and generalization beyond fixed discretizations. Recent neural operator methods have demonstrated accuracy and speedups across diverse PDE settings \cite{azizzadenesheli2024neural}. Collectively, they are establishing themselves as a promising foundation for real-time surrogates in complex simulations.

Two representative families of the various neural operator designs are the Fourier Neural Operator (FNO) and the Deep Operator Network (DeepONet). The FNO encodes inputs in the spectral domain via FFTs, achieving discretization-invariant learning and strong generalization across resolutions and geometries \cite{li2020fourier}. \textcolor{black}{While powerful, FNO’s reliance on global Fourier modes limits its ability to capture local discontinuities. To address these limitations, a series of architectural variants hybridize global spectral layers with localized or hierarchical modules to alleviate spectral bias and capture multiscale structure. Conv-FNO integrates U-Net features; AM-FNO amortizes Fourier kernel parameterization; and the Wavelet Neural Operator (WNO) replaces Fourier transforms with wavelets to enhance spatial localization \cite{liu2025enhancing,xiao2024amortized,tripura2022wavelet}. More recent designs further incorporate hierarchical attention and scale-adaptive receptive fields \cite{liu2024mitigating,qin2024toward}, including U-FNO \cite{wen2022u}, factorized FNO (F-FNO) \cite{tranfactorized}, hierarchical attentive FNO \cite{zhao2023enhancing}, and hierarchical transformers (HT-Net) \cite{liu2022ht}. Complementary lines distribute modeling capacity or long-range information exchange: D2NO partitions the input function space with per-partition subnetworks and a global aggregator \cite{zhang2024d2no}, while attention-based operators such as GNOT and ViTO employ (cross-)attention to handle irregular discretizations and multi-input settings efficiently \cite{hao2023gnot,ovadia2024vito}. Taken together, these ablation-driven architectures reveal a central design trade-off between global parameter sharing and local specialization, which directly motivates our coupling-aware study of DeepONets.}

\textcolor{black}{Within this context,} DeepONet provides a complementary operator-learning architecture~\cite{lu2021learning}. It is grounded in the universal approximation theorem for nonlinear operators \cite{chen1995universal}, which establishes that a branch–trunk network can approximate any continuous operator to arbitrary accuracy. In practice, the branch network encodes input functions, and the trunk network encodes spatial or spatiotemporal coordinates. Their outputs are combined to predict solution fields at arbitrary query points, enabling flexible resolution and geometry handling. DeepONet has been extended with physics-informed constraints (PI-DeepONet), separable structures (Sep-PI-DeepONet), and recurrent modules for temporal dynamics \cite{wang2021learning, mandl2025separable, he2024sequential}. These extensions have powered applications in nonlinear solid mechanics, fracture mechanics, aerodynamics, seismology, fluid flow, and heat transfer \cite{koric2024deep, goswami2022physics, zhao2023learning, xu2023training, haghighat2024deeponet, kobayashi2024improved, kobayashi2025proxies, kobayashi2024deep, hossain2025virtual, sahin2024deep, koric2023data}. Performance improvements have also been achieved through orthonormalized two-stage training \cite{lee2024training}, multifidelity frameworks \cite{howard2023multifidelity}, and enriched moving-solution operators \cite{haghighat2024deeponet}. Modular designs such as DeepM\&Mnet \cite{cai2021deepm, mao2021deepm} and Fourier-Multiple-Input Operator Network (MIONet) \cite{jiang2024fourier} highlight how DeepONet can be adapted for multiphysics systems while reducing parameter count.
Building on these two representative frameworks, further innovations have broadened the scope of operator learning. Physics-informed neural operators (PINO) enforce governing equations during training \cite{li2024physics}, Physics-Informed Parallel Neural Operators (PIPNO) extend this to data-free regimes \cite{yuan2025high}, and Codomain Attention Operators (CoDA-NO) improve generalization across coupled physics through output-space attention \cite{rahman2024pretraining}. Geometry-aware approaches embed diffeomorphic transformations to enable variable-domain learning \cite{zhao2025diffeomorphism}. Together, these developments underscore the rapid diversification of operator architectures and their potential for complex scientific computing.

Neural operators have already demonstrated real-world utility in high-consequence domains. FNO-based surrogates have enabled closed-loop control in additive manufacturing \cite{liu2024deep}, while operator-based digital twins \cite{kobayashi2024deep, kobayashi2024explainable, daniell2025digital,roy2026adversarial,howes2026graph,yoo2025cross} have accelerated reactor monitoring, anomaly detection, and transport prediction in advanced nuclear systems (e.g., \cite{alam2019neutronic,alam2019small1,alam2019small2,ahmed2021numerical}). These successes highlight their scalability, data efficiency, and capability for real-time deployment.

However, extending these methods to strongly coupled multiphysics systems remains a central challenge. Many practical systems involve nonlinear, interdependent processes (e.g., thermo-mechanical, electro-thermal, or fluid–structure coupling) \cite{yang2025multiphysics}. High-fidelity multiphysics simulations are computationally expensive but critical for ensuring performance and safety. Surrogate models such as DeepONet and related architectures could provide orders-of-magnitude speedups \cite{kushwaha2024advanced}. However, whether current operators can capture strong nonlinear inter-field couplings without accumulating errors is unclear. In practice, existing operator networks often struggle due to the complexity of multiphysics interactions and the limited availability of high-resolution coupled-training data \cite{rahman2024pretraining}. No systematic framework has been proposed to test or guide architectural selection based on coupling strength.

This work addresses that gap. We present a systematic topological ablation study of DeepONet and S-DeepONet architectures across problems of increasing physical coupling strength. Specifically, we compare single-branch and multi-branch (MIONet-style) topologies within these two operator families, and demonstrate how branch configuration affects predictive accuracy as coupling strength varies. The findings yield practical architectural insights scoped to DeepONet and S-DeepONet topologies. We do not claim architectural universality beyond these model families.

\section*{Results}
DeepONet learns nonlinear operators that transform between functional spaces, including those that denote integral operators, partial differential equations (PDEs), or dynamical systems. In a functional space $U$, $u \in U$ represents the input parameters (i.e., input functions), and $s \in S$ refers to an unknown PDE solution in the functional space $S$.

We assume a single solution $s = s(u)$ in $S$ for every $u \in U$, which is also subject to the boundary conditions. As a result, the mapping solution operator $G: U \rightarrow S$, evaluated at continuous spatiotemporal coordinates $\xi$ can be defined as: 

\begin{equation}
    G(u)(\xi) = s(u)(\xi)
\label{eq:8}
\end{equation}


In its original form, DeepONet architecture\cite{lu2021learning} consisted of two forward fully connected neural networks (FNNs): a branch net, that encodes the input function $u$ at fixed sensor locations, to $B_h$ output, and a trunk net that encodes the n spatiotemporal coordinates, where the output solution is defined, producing the output $T_{nh}$. Dimension index $h$ represents the hidden output dimension of the branch and trunk networks. DeepONet considers both $u$ and $\xi$ and predicts a solution by combining their encoded outputs via a dot product over the hidden dimension $h$, including bias $\beta$. It then compares these outputs to the target solutions (labels) provided by classical numerical analysis and calculates the loss function, which represents the prediction error. Subsequently, during the backpropagation process, the gradient of the loss function with respect to the trainable parameters $\theta$ in both networks is computed, and the optimizer reduces the loss value by modifying the trainable parameters. After completing a sufficient number of feedforward and backpropagation iterations, referred to as epochs, DeepONet can predict solutions across the entire domain, also known as solution fields. 


In cases of multiple input functions, one can read and encode them jointly in a single branch network, or each input function can have its own branch network that encodes it independently. The encoded outputs from multiple branches are joined by the element-by-element tensor (Hadamard) product before the dot product with trunk output, as first proposed in MIONet architecture \cite{jin2022mionet}. In that work, single-physics PDE examples with multiple input functions were trained, and the advantage of MIONet in terms of prediction accuracy was demonstrated over single-branch DeepONet. However, that work did not consider multiphysics cases where physics phenomena are coupled and tightly dependent on each other, including the path-dependent behavior, where the final state of such a system doesn’t just depend on where it ends up, but also on how it got there through the order of elements in sequential input functions.

\begin{figure}[!htbp]
    \centering
    \includegraphics[width=1.0\textwidth]{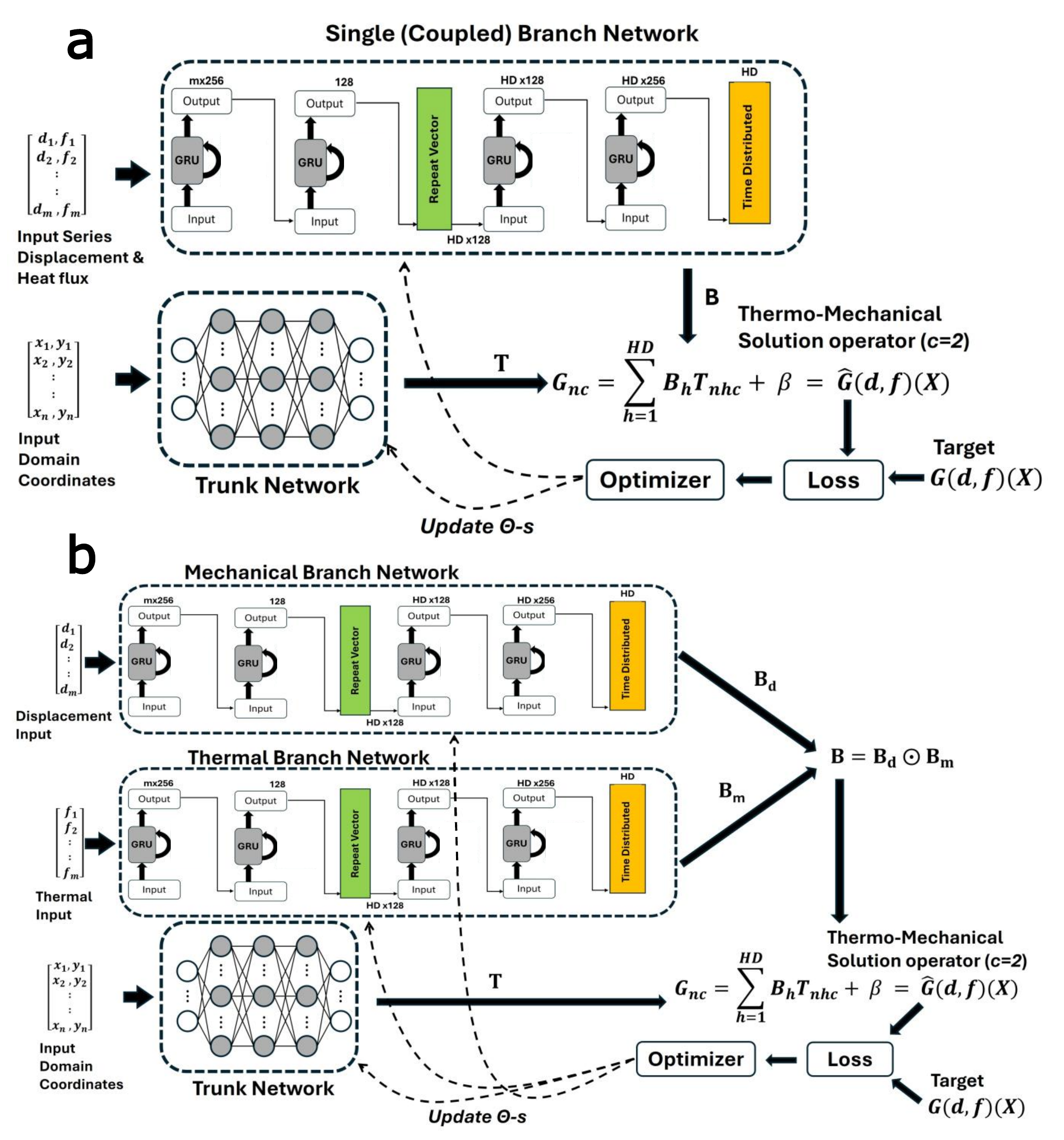}
    \caption{\textbf{Two sequential deep operator network (S-DeepONet) architectural variants compared in this study.} 
    \textbf{(a)}, Single-branch (coupled) design. A gated recurrent unit (GRU)-based encoder--decoder processes the sequential inputs jointly, producing branch output $B_h$ that is combined with trunk encoding $T_{nhc}$, where m is the input time series length and HD is the hidden dimension.
    \textbf{(b)},Two-branch (multiple-input operator network, MIONet-style) design. Independent GRU branches produce $B_d$ and $B_m$, which are merged via the Hadamard product before interaction with trunk encoding $T_{nhc}$. 
    The symbol $\theta$ denotes the trainable parameters contained in the model.}
    \label{fig:sdon_arch}
\end{figure}

In multiphysics problems, the underlying physical processes are often driven by sequential input functions, and the characteristics of the governing physics, such as path dependency in plastic deformation, necessitate a complete load history to accurately determine the system's final state. Since the original DeepONet architecture with a feedforward fully connected neural network (FNN) in the branch does not capture the sequential nature of its input, the authors recently introduced the sequential DeepONet (S-DeepONet) \cite{he2024sequential}, which employs a gated recurrent unit (GRU) based encoder-decoder model in the branch network, replacing the FNN for better encoding sequential input functions. GRU are more suitable for sequential learning than standard recursive neural networks because they address vanishing gradient issues with gating mechanisms, thereby enabling the learning of long-term dependencies. The S-DeepONet model demonstrated enhanced predictive accuracy \cite{he2024sequential} compared to the original DeepONet, albeit with an extended training duration. Fig.~\ref{fig:sdon_arch}(a) shows the architecture with a single-branch that encodes both inputs $d$ and $f$ together in a coupled fashion through a series of GRU cells arranged in an encoder-decoder fashion, resulting in encoded branch output $B_h$ with hidden dimension size HD (or $h$). Similarly, the trunk network takes $n$ input domain coordinates and encodes them into its multidimensional output $T_{nhc}$, where $c$ represents the number of output solution fields predicted simultaneously, i.e., $c=2$, corresponding to temperature and stress. The encoded outputs from the branch and trunk are joined via a dot product to form the final S-DeepONet prediction $G_{nc}$ via a dot product along the hidden dimension, including bias $\beta$. 

Inspired by MIONet \cite{jin2022mionet}, which showed increased prediction accuracy for encoding multiple inputs in single-physics problems, a variant of S-DeepONet that encodes thermal and mechanical inputs in separate encoder-decoder GRU thermal and mechanical branches and then joins their outputs $B_d$ and $B_m$ via Hadamard tensor product, denoted by $\odot$, is devised and represented in Fig.~\ref{fig:sdon_arch}(b). Identical S-DeepONet architectures were trained on the thermo-electrical multiphysics example, using the heat source and charge density input sequences to branch, while the spatiotemporal $x,t$ coordinate pairs were input into the trunk.

These architectures, including the classical DeepONet, multi-branch MIONet, and S-DeepONet (in both single- and multi-branch forms), serve as the comparative baselines for the benchmark problems presented in the following subsections.

\subsection*{Single Physics Reaction-Diffusion Model}
\label{sec:single_phys_reaction_diffusion}
The first benchmark problem involves a single-physics reaction–diffusion PDE, representative of transport-dominated processes such as thermal conduction, chemical diffusion, and drug dispersion in tissue:  

\begin{equation}
\frac{\partial u}{\partial t} = 0.01 \nabla^2 u + u_0(x) - k(x) u,
\label{eq:1}
\end{equation}

\textcolor{black}{where $u(x,t)$ denotes the state field (e.g., concentration or temperature) at spatial location $x$ and time $t$, $u_0(x)$ is a heterogeneous source term of the corresponding state field,} and $k(x)$ is a spatially varying reaction coefficient. Both functions were generated as Gaussian random fields using \texttt{gstools}~\cite{muller2022gstools}. Finite-element simulations (FEM) were performed with \texttt{FEniCSx}~\cite{baratta2023dolfinx} on a one-dimensional domain discretized with 127 quadratic elements (255 nodes) over 101 time steps, producing a dataset of 10,000 full spatiotemporal solutions $u(x,t)$, of which 8,000 were used for training and 2,000 for testing.  

Each sample consists of two input fields $[u_0(x), k(x)]$ defined on the 255-node mesh, giving inputs per sample of shape [255, 2]. The corresponding solution field $u(x,t)$ spans 101 temporal steps and 255 spatial nodes, with outputs per sample of shape [101, 255]. We trained both the single-branch DeepONet, where $[u_0(x), k(x)]$ are jointly encoded into $B_h$, and the multi-branch MIONet variant, where $u_0(x)$ and $k(x)$ are independently encoded into $B_d$ and $B_m$ before merging via Hadamard product.  

Model performance was quantified using the relative $L_2$ error between predicted solutions $S_{pred}$ and finite-element ground truth $S_{FE}$:  

\begin{equation}
L_2 = \frac{\lVert S_{FE} - S_{pred} \rVert_2}{\lVert S_{FE} \rVert_2}.
\label{eq:l2_error}
\end{equation}

\begin{figure}[!htbp]
    \centering
    \includegraphics[width=\textwidth]{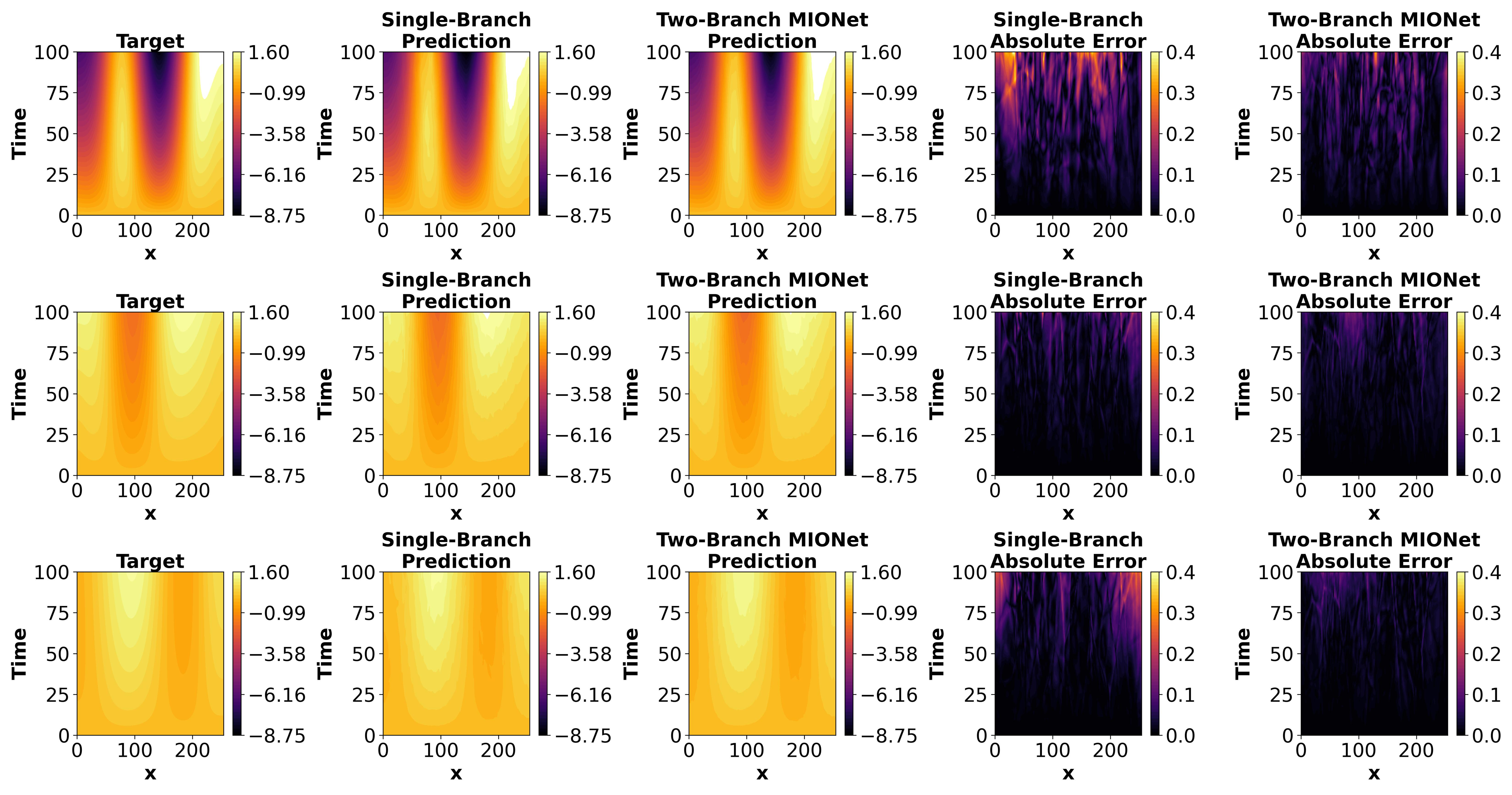}
    \caption{\textbf{Reaction–diffusion benchmark comparing single- and two-branch deep operator network (DeepONet) architectures.} Target finite-element solutions, model predictions, and absolute error contours are shown for representative test samples at the 55th (top), 85th (middle), and 99th (bottom) error percentiles. The multi-branch (multiple-input operator network, MIONet-style) design yields consistently smaller error fields than the single-branch variant. \textcolor{black}{Colorbars are standardized across subfigures: solution panels share a common range, and absolute-error panels share a separate common range.}}
    \label{fig:6}
\end{figure}

On the held-out 20\% test set (never seen during training), the mean L2 error was 3.40\% for the single-branch and 2.55\% for the two-branch MIONet. The architecture is a classical DeepONet with branches equipped with fully connected (FNN) networks, as the input functions $u_0(x)$ and $k(x)$ are of spatial rather than temporal nature and do not induce path-dependent responses. Fig. \ref{fig:6} shows the full spatial-temporal solution field comparison between single-branch and two-branch models for 55th (top row), 85th (middle row), and 99th percentile (bottom row) test samples, with 0\% corresponding to the best test sample characterized by the smallest error, and 99\% corresponding to the worst test sample, characterized by the largest error. 

It can be observed, particularly from the contours of absolute errors, that the MIONet-based DeepONet, when reading and encoding spatial input functions separately, exhibits visibly less error in the domain compared to the single-branch DeepONet. This is entirely in agreement with the observations reported in the original MIONet framework~\cite{jin2022mionet}.

\subsection*{Coupled Electrical Conduction and Thermal Diffusion Model}
\label{sec:electro_thermal}

The first multiphysics use case is the electrical conduction and thermal diffusion PDE system with multiple inputs. The coupled electrical conduction and thermal diffusion model describes the highly coupled interaction between electrical currents and heat flow in a material. \textcolor{black}{These PDEs have essential applications in thermal management in electronics and semiconductor devices \cite{Antoulinakis2016}, batteries and energy storage systems \cite{Kalungi2025}, tissue ablation in biomedical applications \cite{LopezMolina2016}, electroplating \cite{Sun2018}, and welding in manufacturing \cite{Feulvarch2004}.} The normalized governing equations for the coupled model are presented as follows:

\begin{equation}
\left\{
\begin{aligned}
\frac{\partial T(x,t)}{\partial t} &= \nabla \cdot (k \nabla T(x,t)) + Q_{\text{ext}}(t) + Q_e(x,t) \\
\frac{\partial^2 \phi(x,t)}{\partial t^2} &= \nabla \cdot (\gamma \nabla \phi(x,t)) + \rho_e(t)
\end{aligned}
\right.
\label{eq:2}
\end{equation}
\textcolor{black}{where $T(x,t)$ denotes the temperature field and $\phi(x,t)$ the electric potential field. Here, $k=0.116$ denotes the thermal conductivity,  $Q_{\text{ext}}(t)$ represents the heat source term, and $Q_e(x,t) = \gamma |\nabla \phi(x,t)|^2$ is the heat generated by electrical currents. The solution $\phi(x,t)$ is the electric potential, while $\gamma$ refers to the electrical conductivity of the material, defined as $\gamma = \frac{1}{1 + \beta T(x,t)}$ for metals, where $\beta = 3.9$ is the temperature coefficient of resistivity. Additionally, $\rho_e(t)$ is the free charge density, which accounts for any non-zero net charge within the system.}

In this coupling case, the time-dependent input functions $Q_{\text{ext}}(t)$ and $\rho_e(t)$ were randomly generated following a Gaussian distribution using the gstools \cite{muller2022gstools} package. Finite element simulations were then conducted to solve for $T$ and $\phi$  within a one-dimensional domain consisting of 127 second-order elements over 101 time steps to generate 5000 samples, using the FEniCSx \cite{baratta2023dolfinx}. This dataset captures a wide range of temporal input conditions, enabling trained neural network models to generalize effectively across unseen input scenarios and the corresponding diverse scenarios of thermal and electrical solution fields, as well as their interactions.

\textcolor{black}{Besides the coupled solutions, uncoupled solutions were also prepared to represent a less challenging case for predicting the individual physical fields. In this configuration, a single temperature field is passed into the electrical conduction equation to predict the electric potential distribution, which depends only on $\rho_e(t)$. Similarly, a single electric potential field is passed into the heat diffusion equation to predict the temperature, which depends only on the external heat source $Q_{\text{ext}}(t)$.}

\textcolor{black}{The S-DeepONet architecture was applied in two configurations: a single-branch design, in which $[Q_{\text{ext}}(t), \rho_e(t)]$ were jointly encoded into $B_h^{\mathrm{GRU}}$, and a two-branch MIONet-style extension, in which $Q_{\text{ext}}(t)$ and $\rho_e(t)$ were independently encoded into $B_d$ and $B_m$ before being merged via the Hadamard product. Model performance was evaluated using the relative $L_2$ error.}

\begin{figure}[htbp]
    \centering
    \includegraphics[width=0.95\textwidth]{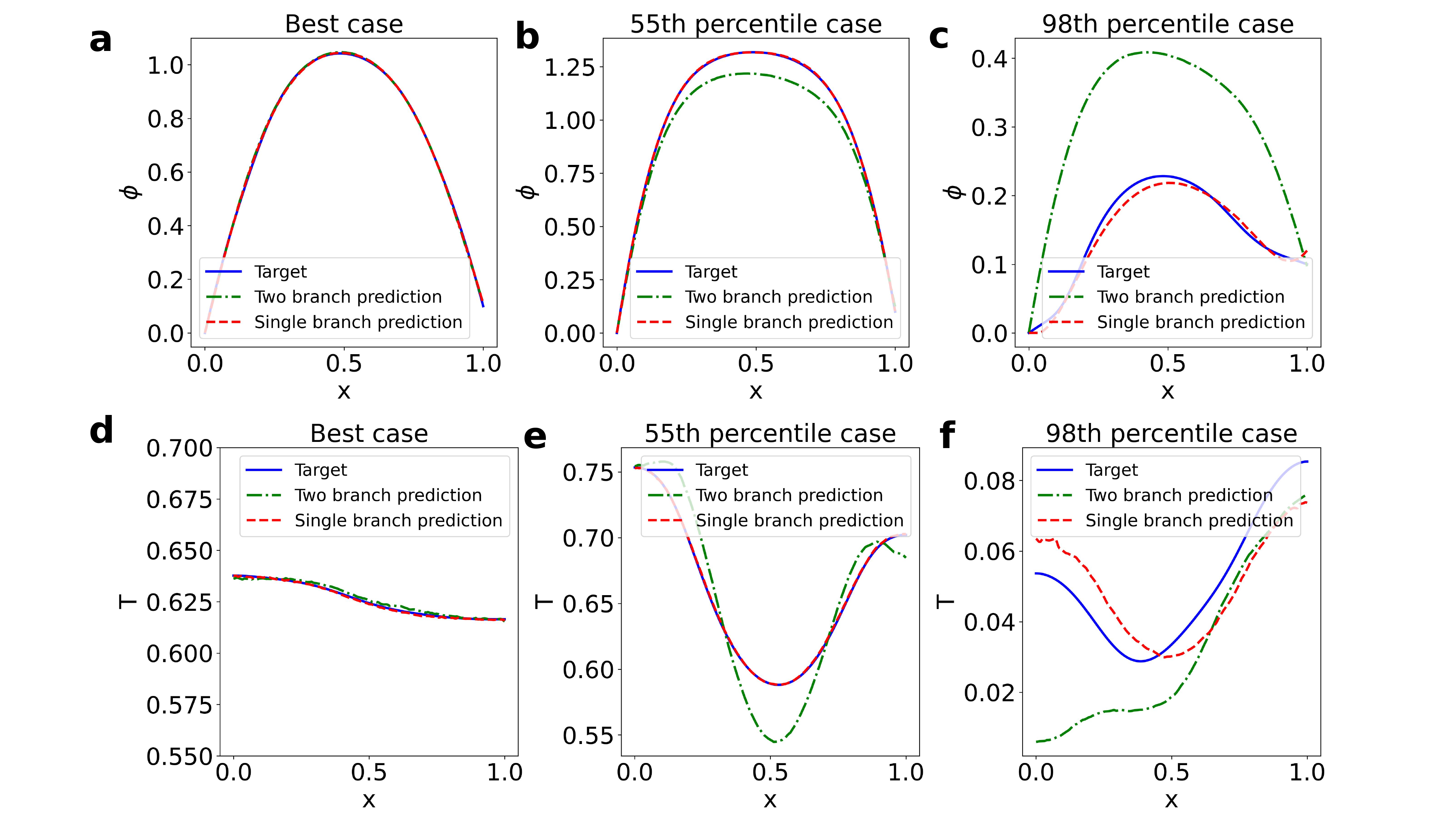}
    \caption{
    \textbf{Predictions of electrical potential and temperature in the coupled electro-thermal benchmark.} 
    \textbf{(a)--(c)}, Electrical potential $\phi$ for the best, 55th percentile, and 98th percentile test cases. 
    \textbf{(d)--(f)}, Temperature $T$ for the same percentile cases. 
    Finite-element method (FEM) reference solutions (Target) are compared with predictions from the single-branch and two-branch sequential deep operator network (S-DeepONet) models. The single-branch model remains close to the reference across all percentiles, whereas the two-branch model exhibits larger deviations at mid- and high-error cases.}
\label{fig:electro_thermal_profiles}
\end{figure}

Fig.~\ref{fig:electro_thermal_profiles}(a)--(c) compares the predicted and target electrical potential values. At the same time, Fig.~\ref{fig:electro_thermal_profiles}(d)--(f) presents the temperature results at the final time step along the domain direction for coupled multiphysics data using both the single-branch and two-branch (MIONet) S-DeepONet models. The single-branch S-DeepONet implementation yields temperature and electrical potential solutions that align more closely with the FEM solution (target) compared to the two-branch S-DeepONet.

\begin{figure}[htbp]
    \centering
    \includegraphics[width=0.85\textwidth]{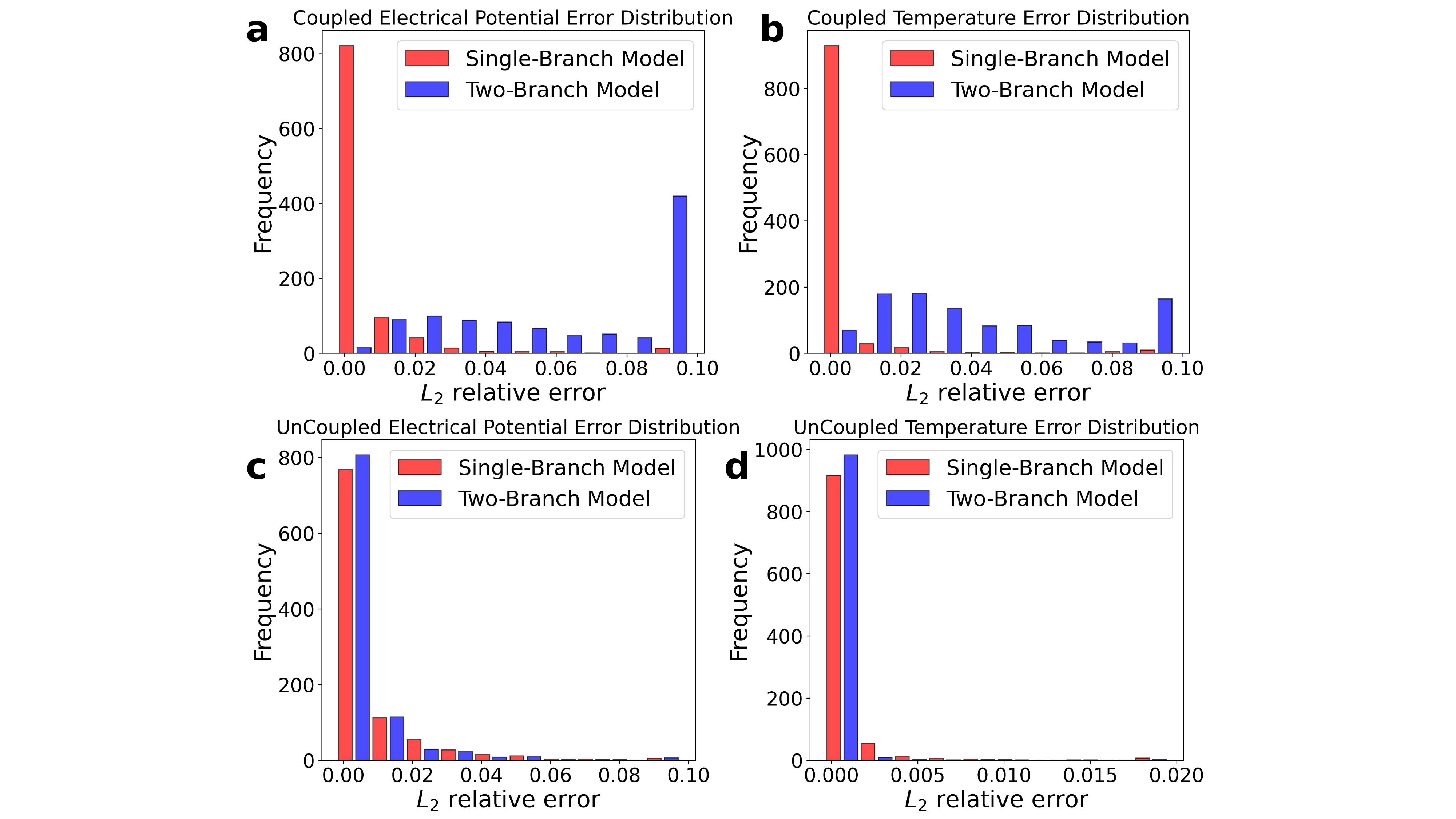}
    \caption{\textbf{Test-error distributions for the electro-thermal benchmark under coupled and uncoupled regimes.} 
    Coupled setting \textbf{(a), (b)}: histograms of $L_2$ relative error for electrical potential ($\phi$) and temperature ($T$). The single-branch model concentrates errors near zero, whereas the two-branch model yields long-tailed distributions with errors exceeding 10\%, particularly for electrical potential. 
    Uncoupled setting \textbf{(c), (d)}: corresponding histograms for electrical potential ($\phi$) and temperature ($T$). The performance gap narrows and the two-branch model slightly outperforms the single-branch, consistent with its advantage when handling independent input fields.}
\label{fig:electro_thermal_hist}
\end{figure}

Additionally, the histogram of test error distributions in Fig.~\ref{fig:electro_thermal_hist}(a)--(b) highlights the accuracy of the single-branch implementation over the two-branch implementation, particularly for the electrical potential. The two-branch predictions for electrical potential (represented by blue columns) exhibit a long tail, indicating larger error areas.

Fig.~\ref{fig:electro_thermal_hist}(c)--(d) indicates that, when the fields are uncoupled, the performance gap narrows and the two-branch model is slightly favored. This accords with the intuition and our single-physics benchmark that independent inputs benefit from separate encoders, while tightly coupled multiphysics demands a shared-parameter latent.

\textcolor{black}{Table~\ref{tab:combined_error_potential_temperature} summarizes the $L_2$ relative error results, obtained by repeating training and evaluation five times using different random seeds while keeping the model architecture, hyperparameters, and training schedules fixed. The single-branch S-DeepONet, which reads both inputs together and encodes them with shared (coupled) trainable parameters, is more accurate with coupled multiphysics (thermo-electrical) analysis data than S-DeepONet MIONet with two branches, and this is particularly the case for electrical potential, where MIONet-based S-DeepONet has over 15\% error. In contrast, the single-branch S-DeepONet has slightly over 1\% error.}

On the other hand, uncoupled data represents a substantially less challenging task for prediction, and both architectures perform comparably well. Since each physics depends on its input only, the two-branch (MIONet-based) S-DeepONet model is slightly more accurate by reading and encoding each input separately and joining them via the Hadamard tensor product.



\begin{table}[!htbp]
\caption{Mean $L_2$ relative error for temperature and electrical potential (mean $\pm$ standard deviation over five independent runs). Bold values indicate the lower error between single-branch and two-branch architectures for each physics regime and output field.}
\centering
\begin{tabular}{@{}lcccc@{}}
\toprule
 & \multicolumn{2}{c}{\textbf{Temperature (\%)}} & \multicolumn{2}{c}{\textbf{Electrical Potential (\%)}} \\
\cmidrule(lr){2-3} \cmidrule(l){4-5}
\textbf{Physics} & \textbf{Single-Branch} & \textbf{Two-Branch} & \textbf{Single-Branch} & \textbf{Two-Branch} \\
\midrule
Coupled   & $\boldsymbol{0.53 \pm 0.04}$   & $3.81 \pm 0.86$ & $\boldsymbol{0.76 \pm 0.18}$ & $4.23 \pm 1.00$ \\
Uncoupled & $0.081 \pm 0.031$          & $\boldsymbol{0.070 \pm 0.021}$ & $0.55 \pm 0.33$          & $\boldsymbol{0.37 \pm 0.28}$ \\
\bottomrule
\end{tabular}
\label{tab:combined_error_potential_temperature}
\end{table}

\subsection*{Thermo-Mechanical Model of Steel Solidification}
\label{sec:thermo_mech}

\textcolor{black}{Although advancements in metal-based additive manufacturing are progressing steadily, the materials processing domain remains predominantly governed by conventional techniques, such as ingot, foundry, and continuous casting. Continuous casting accounts for more than 95\% of global steel production \cite{thomas2018review,Louhenkilpi2014}.} Experiments are constrained by the severe environment of molten steel and the numerous process variables that influence its intricate multiphysics phenomena. The progress of these established manufacturing processes mostly depends on the enhanced quantitative insights obtained from advanced multiphysics numerical models. \textcolor{black}{Major advancements in computing technology and numerical techniques over the past 30 years have enabled more realistic and precise multiphysics simulations of steel solidification processes on high-performance computing systems, utilizing specialized software in predominantly offline academic research environments \cite{koric2010multiphysics}.} However, optimization, designs, and real-time predictions for online controls of these processes are still primarily impossible by classical modeling methods.

The thermal field is governed by the enthalpy form of the heat equation,
\begin{equation}
\rho \,\frac{\partial H}{\partial t}
=
\boldsymbol{\nabla} \cdot \bigl(k\,\boldsymbol{\nabla} T\bigr),
\label{eq:3}
\end{equation}
in which $H(T)$ represents specific enthalpy incorporating both sensible heat associated with temperature change and latent heat associated with phase-transformation during solidification and melting. $k$ and $\rho$ denote the
thermal conductivity and mass density, respectively. Because the casting time scale is much longer than any acoustic
transit time across the slice, momentum transients can be safely neglected and the mechanical balance reduces to its
quasi-static form,
\begin{equation}
\nabla \cdot \boldsymbol{\sigma}(\mathbf{x}) + \mathbf{b} = \mathbf{0},
\label{eq:4}
\end{equation}
with Cauchy stress $\boldsymbol{\sigma}$ and body force density $\mathbf{b}$. The total strain rate decomposes as
\begin{equation}
\dot{\boldsymbol{\varepsilon}} 
= \dot{\boldsymbol{\varepsilon}}_{\text{el}} 
+ \dot{\boldsymbol{\varepsilon}}_{\text{ie}} 
+ \dot{\boldsymbol{\varepsilon}}_{\text{th}},
\label{eq:5}
\end{equation}
and the constitutive relation reads
\begin{equation}
\dot{\boldsymbol{\sigma}} 
= \mathbf{D} : \bigl( \dot{\boldsymbol{\varepsilon}} 
- \alpha\,\mathbf{I}\,\dot{T} 
- \dot{\boldsymbol{\varepsilon}}_{\mathrm{ie}} \bigr),
\label{eq:6}
\end{equation}
\textcolor{black}{where $\mathbf{D}$ contains temperature-dependent elastic constants, $\alpha$ is the (temperature-dependent) thermal expansion coefficient, and $\mathbf{I}$ is the identity. High-temperature inelasticity in austenite is modeled by the Kozlowski visco-plastic law \cite{kozlowski1992simple},}
\begin{equation}
\dot{\varepsilon}_{\mathrm{ie}}\;[\mathrm{s}^{-1}]
\;=\;
f_{c}\,
\Bigl(\bar{\sigma}\;[\mathrm{MPa}]\;-\;f_{1}\,\bar{\varepsilon}_{\mathrm{ie}}\;\bigl|\dot{\varepsilon}_{\mathrm{ie}}\bigr|^{\,f_{2}-1}\Bigr)^{\,f_{3}}
\exp\Bigl(-\frac{Q}{T\;[\mathrm{K}]}\Bigr),
\label{eq:7}
\end{equation}
where
\[
\begin{aligned}
Q &= 44{,}465,\\
f_{1} &= 130.5 \;-\; 5.128 \times 10^{-3}\,T\;[\mathrm{K}],\\
f_{2} &= -0.6289 \;+\; 1.114 \times 10^{-3}\,T\;[\mathrm{K}],\\
f_{3} &= 8.132 \;-\; 1.54 \times 10^{-3}\,T\;[\mathrm{K}],\\
f_{c} &= 46{,}550 \;+\; 71{,}400\,(\%\mathrm{C}) \;+\; 12{,}000\,(\%\mathrm{C})^{2}.
\end{aligned}
\]

\textcolor{black}{$Q$ is an activation energy constant in ($K$) defined as activation energy in ($Jmol^{-1}$) over gas constant ($Jmol^{-1}K^{-1}$), $\bar{\sigma}$ is Von Mises effective stress (MPa), empirical functions  $f_1$, $f_2$, $f_3$, and $f_c$ depend on absolute temperature (K), and \%C is carbon content (weight percent) representing steel grade (composition).} Another constitutive model, the so-called Zhu power law model \cite{zhu1996coupled} was devised to simulate the delta ferrite phase with a relatively higher creep rate and weaker than the austenite phase and is applied in the solid whenever the delta-ferrite's volume fraction is more than 10\%, to approximate the dominating influence of the very high-creep rates in the delta-ferrite phase of mixed-phase structures on the net mechanical behavior. This work applies the elastic-perfectly-plastic constitutive model with small yield stress above the solidus temperature $T_{\text{sol}}$ to enforce negligible strength in those volatile zones. The highly nonlinear constitutive visco-plastic austenite or delta-ferrite models are efficiently integrated at the integration points in the UMAT subroutine \cite{koric2006efficient} and linked with Abaqus \cite{abaqus2022} implicit finite element analysis (FEA) software.

\begin{figure}[htbp]
    \centering
    \includegraphics[width=0.95\textwidth]{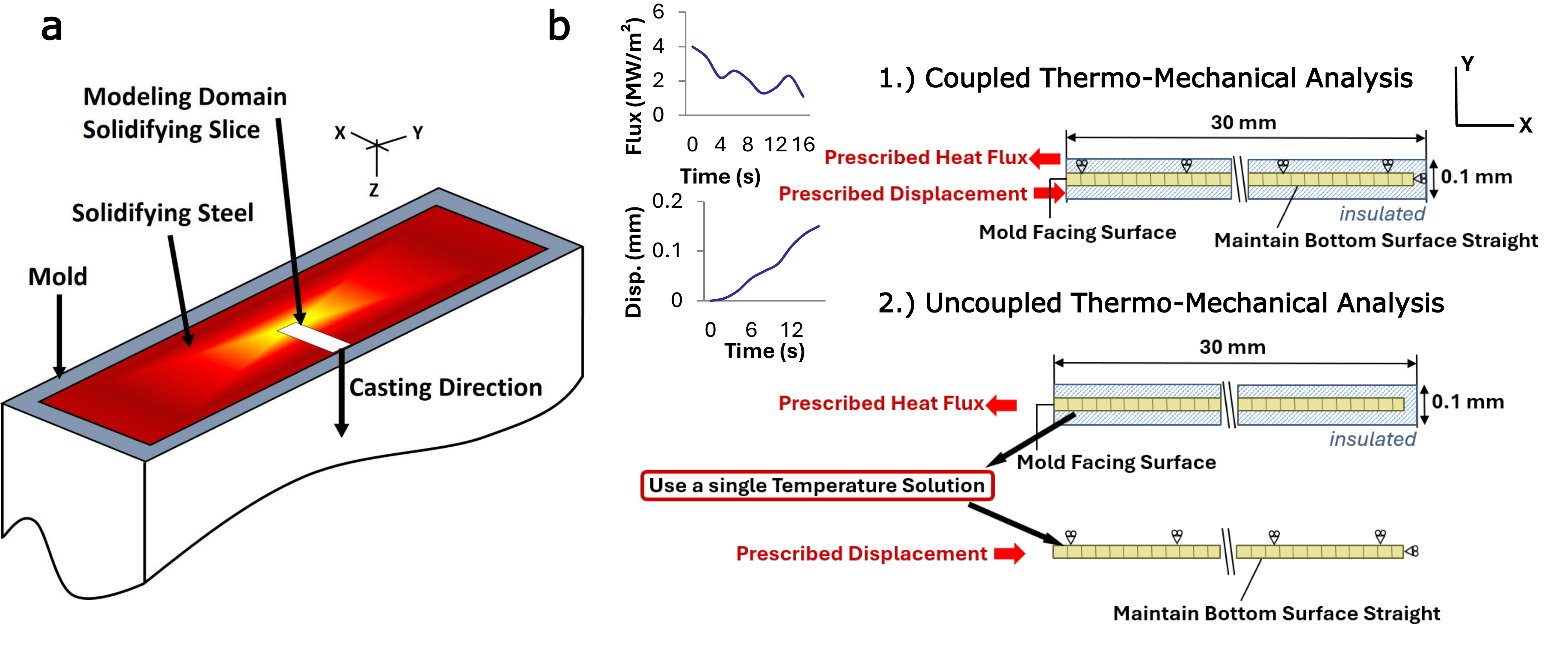}
    \caption{
    \textbf{Thermo--mechanical solidification context and boundary value problems for data generation.} 
    \textbf{(a)}, Modeling slice within the continuous-casting context. A representative section of the slab moves through the mold in the casting direction; the highlighted slice denotes the modeling domain used for data generation.~\cite{he2024sequential}. \textbf{(b)}, Boundary value problems for the coupled and uncoupled analyses. In the coupled case (top), simultaneous time-varying heat flux and normal displacement are prescribed on the mold-facing surface under generalized plane strain, with the bottom edge constrained to remain straight. In the uncoupled case (bottom), a single temperature solution is paired with prescribed displacement, isolating the mechanical response to mold taper without thermo--mechanical feedback.}
\label{fig:solid_setup}
\end{figure}

The phase fraction and temperature-dependent material properties calculations for the low carbon steel grade chosen for this work with 0.09 wt\%C, having solidus and liquidus temperatures $T_{\text{sol}}=1480.0 ^\circ$C and $T_{\text{liq}} = 1520.7 ^{\circ}$C are also an integral part of the UMAT subroutine, and more details about this robust multiphysics model can be found in \cite{zappulla2020multiphysics}. When subjected to thermal loading, generalized plane strain conditions can effectively restore a complete three-dimensional stress state in long objects, such as the continuous caster shown in Fig.~\ref{fig:solid_setup}(a), which has a large length and width. The slice domain moves down the mold at casting velocity in a Lagrangian frame of reference.  Figure.~\ref{fig:solid_setup}b (top) summarizes the coupled temperature and mechanical conditions employed in addressing the associated boundary value problems (BVP) to produce training and testing data. The generalized plane strain condition in the axial ($z$-direction) is represented by a single row of 300 interconnected thermo-mechanical generalized plane strain 2D elements comprising 602 nodes. Additionally, a second generalized plane strain condition was applied at the lower boundary of the domain by enforcing uniform strain in the y-direction on the vertical displacements of all nodes. The thermal and mechanical boundary conditions of the solidifying slice domain are determined by time-dependent heat flux profiles emanating from the chilled surface on the left side of the domain, as well as their displacement resulting from mold taper and other thermo-mechanical interactions with the mold. 

The research conducted at the steel plant shows that the displacement profile is upwardly trending due to mold taper. In contrast, the thermal flux generally decreases due to transient heat-transfer cooling. A radial basis interpolation using a Gaussian function is employed to interpolate multiple temporal points randomly selected within expected profile ranges, accommodating fluctuations and noise observed in actual flux and displacement profiles due to abrupt changes in contact conditions and interfacial heat transfer between the mold and steel surfaces. They cover all potential thermo-mechanical boundary condition scenarios in which the solidifying shell would encounter at its cooled surface as it descends the caster. \textcolor{black}{Besides data from fully coupled analysis, data samples were generated from uncoupled analysis whose BC-s were depicted Figure ~\ref{fig:solid_setup}b (bottom), where the single temperature solution was passed into a separate mechanical analysis, thus providing us an opportunity to study further how S-DeepONet learns stress solution for uncoupled analysis where stress solution depends on its mechanical (displacement) input only. Although simplified and less accurate, such an uncoupled analysis has some practical significance, as it allows for the independent study of the mechanical effects of the mold’s thermal distortion and taper on a solidifying domain, given a fixed heat flux profile and thus a fixed temperature solution, for the purpose of residual stress generation.}

Finite-element simulations employ a single row of 300 generalized plane-strain thermo-mechanical 2D elements (602 nodes) over 101 time steps. The coupled dataset comprises 5{,}494 samples (4{,}395 for training and 1{,}099 for testing). For each sample, the inputs are sequential histories $\mathbf{u}(t) = [q(t), d_n(t)]$ with shape [101, 2]. The outputs are paired fields, temperature and an effective stress measure (e.g., von Mises), evaluated on the mesh, with output per sample at the final time step of shape [602, 2]. An additional uncoupled dataset contains 3{,}000 samples (2{,}400 for training and 600 for testing), constructed as shown in Figure~\ref{fig:solid_setup}b (bottom).

We compare two S-DeepONet configurations using the notation introduced earlier. In the \emph{single-branch} design, the sequential inputs are jointly encoded into $B_h^{\mathrm{GRU}}$ and combined with the trunk $T_{nhc}$ to produce
\begin{equation}
G_{nc} = B_h^{\mathrm{GRU}} \cdot T_{nhc} + \beta, \quad c=2.
\end{equation}
In the \emph{two-branch (MIONet-style)} design (Fig.~\ref{fig:sdon_arch}(b)), $q(t)$ and $d_n(t)$ are encoded independently into $B_d$ and $B_m$ and merged via the Hadamard product before interacting with the trunk,
\begin{equation}
B_h = B_d \odot B_m, \qquad G_{nc} = B_h \cdot T_{nhc} + \beta.
\end{equation}
This benchmark therefore probes whether a shared latent ($B_h^{\mathrm{GRU}}$) is necessary to learn the joint thermo–mechanical history implied by Eqs.~\eqref{eq:3}–\eqref{eq:7}, versus a factorized latent ($B_d \odot B_m$) that assumes separable encodings.

Owing to the extensive range of prescribed thermal and displacement histories imposed as boundary conditions on the chilled edge, steel solidification in the continuous caster is periodically impeded, resulting in the domain predominantly remaining in the liquid and mushy zones, characterized by negligible strength and minimal, \textcolor{black}{nearly constant stress magnitudes. These instances are not significant in stress-based casting failure forecasts and are outliers.
Accordingly, Eq.~\ref{eq:mae_tm} reports the mean absolute error (MAE) between the finite element (target) and predicted solution values over all testing samples \(N\). MAE is expressed in physical units and remains numerically stable when the ground-truth stress norm is small, whereas relative errors can be dominated by near-zero denominators in predominantly liquid and mushy zones.
Furthermore, the findings are provided within the 85-90\% test error range for both stress and temperature data to mitigate the impact of near-zero stress outliers.}

\begin{equation}
    \mathrm{MAE} = \frac{1}{N}\sum_{i=1}^{N}\left|S_{FE,i} - S_{\mathrm{pred},i}\right|.
\label{eq:mae_tm}
\end{equation}

\begin{figure}[htbp]
    \centering
    \includegraphics[width=0.90\textwidth]{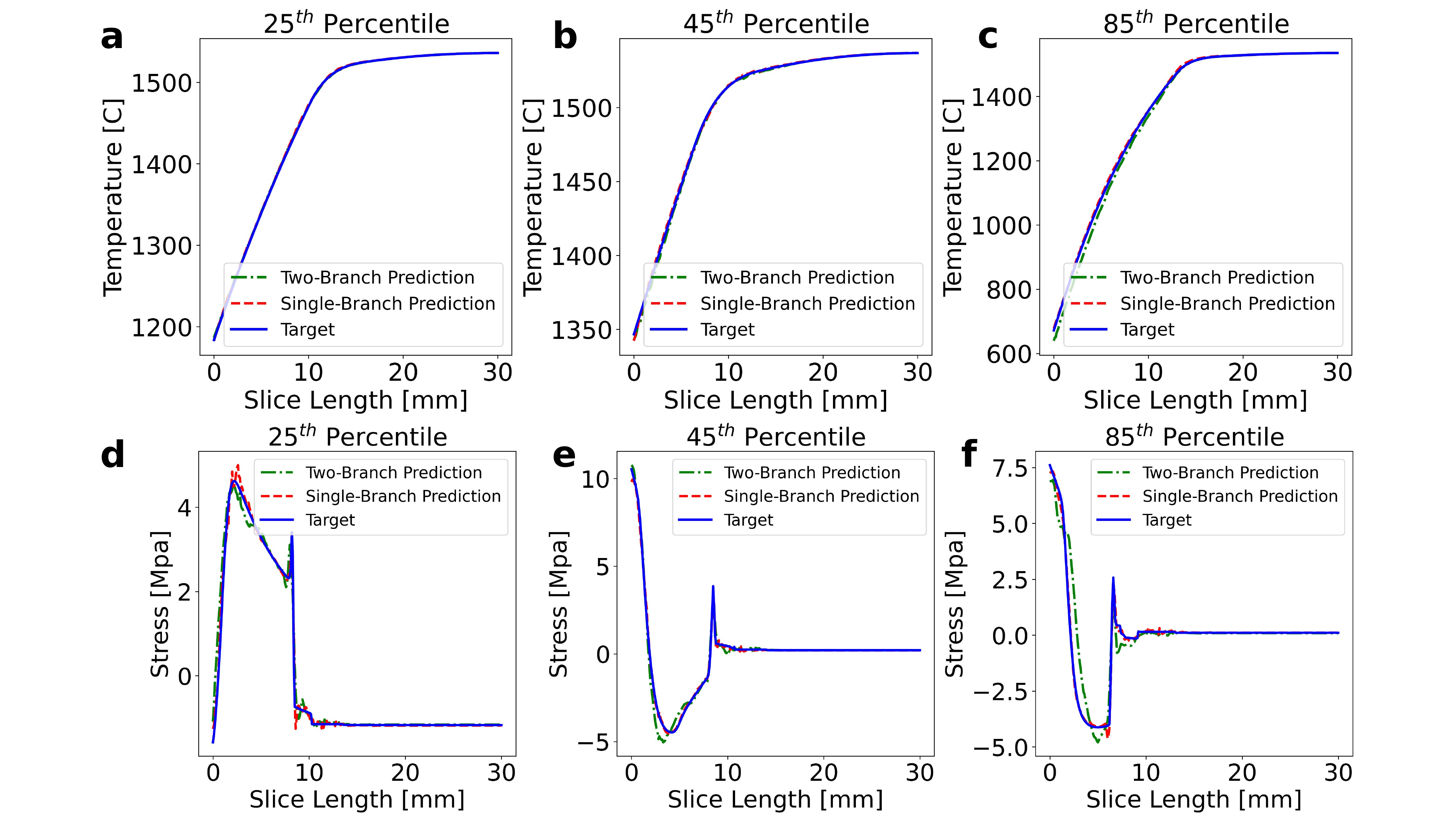}
    \caption{\textbf{Temperature and stress profiles along the slice for the coupled thermo--mechanical benchmark.} 
    \textbf{(a)--(c)}, Temperature along the slice at the 25th (\textbf{a}), 45th (\textbf{b}), and 85th (\textbf{c}) test percentiles. Both single-branch and two-branch sequential deep operator network (S-DeepONet) predictions closely match finite-element method (FEM) targets; even in high-percentile cases, the absolute errors remain small (worst case $\approx 0.65\,^{\circ}\mathrm{C}$). 
    \textbf{(d)--(f)}, Stress along the slice at the same percentiles. The single-branch model better preserves the peak tensile values at the chilled surface and the subsequent stress reversal, whereas the two-branch model deviates at higher percentiles.}
\label{fig:thermo_mech_profiles}
\end{figure}

Fig.~\ref{fig:thermo_mech_profiles}(a)--(c) compares predicted and target temperature values, and Fig.~\ref{fig:thermo_mech_profiles}(d)--(f) shows stress at the mold exit along the slice domain for coupled multiphysics data for single-branch and two-branch (MIONet) S-DeepONet models. Temperature distributions are inherently smoother and less challenging for both S-DeepONet implementations to predict than the stress distribution.  The most unfavorable test case forecasted the complete temperature distribution with an absolute error of 0.65 $\rm ^{\circ}C$, which is highly precise given that the steel temperatures in continuous casting molds range from 1000 $\rm ^{\circ}C$ to 1500 $\rm ^{\circ}C$. Stress distributions are strongly affected by the three constitutive models (austenite, delta-ferrite, and mushy/liquid) determined by the prevailing phase fractions and temperature throughout solidification. Furthermore, stress depends on temperature fluctuations caused by thermal fluxes and the changing displacement history applied to the cooled surface. This results in a highly erratic stress distribution across the slice domain, complicating precise prediction. Despite the evident challenges, both S-DeepONet implementations relatively accurately predicted the stress distribution, including stress reversals with tensile stress at the chilled surface, as illustrated for the $45^{th}$ and $85^{th}$ percentile test samples. These are crucial for forecasting certain failure mechanisms, such as hot tearing. Between the two models, the single-branch S-DeepONet was more accurate with a stress MAE of 0.12 MPa vs. 0.23 MPa for the two-branch S-DeepONet. This is particularly evident in the $85^{th}$ percentile test error, where the two-branch prediction was off from the target, and the single-branch prediction was consistent throughout the solid phase.

\begin{figure}[htbp]
    \centering
    \includegraphics[width=0.90\textwidth]{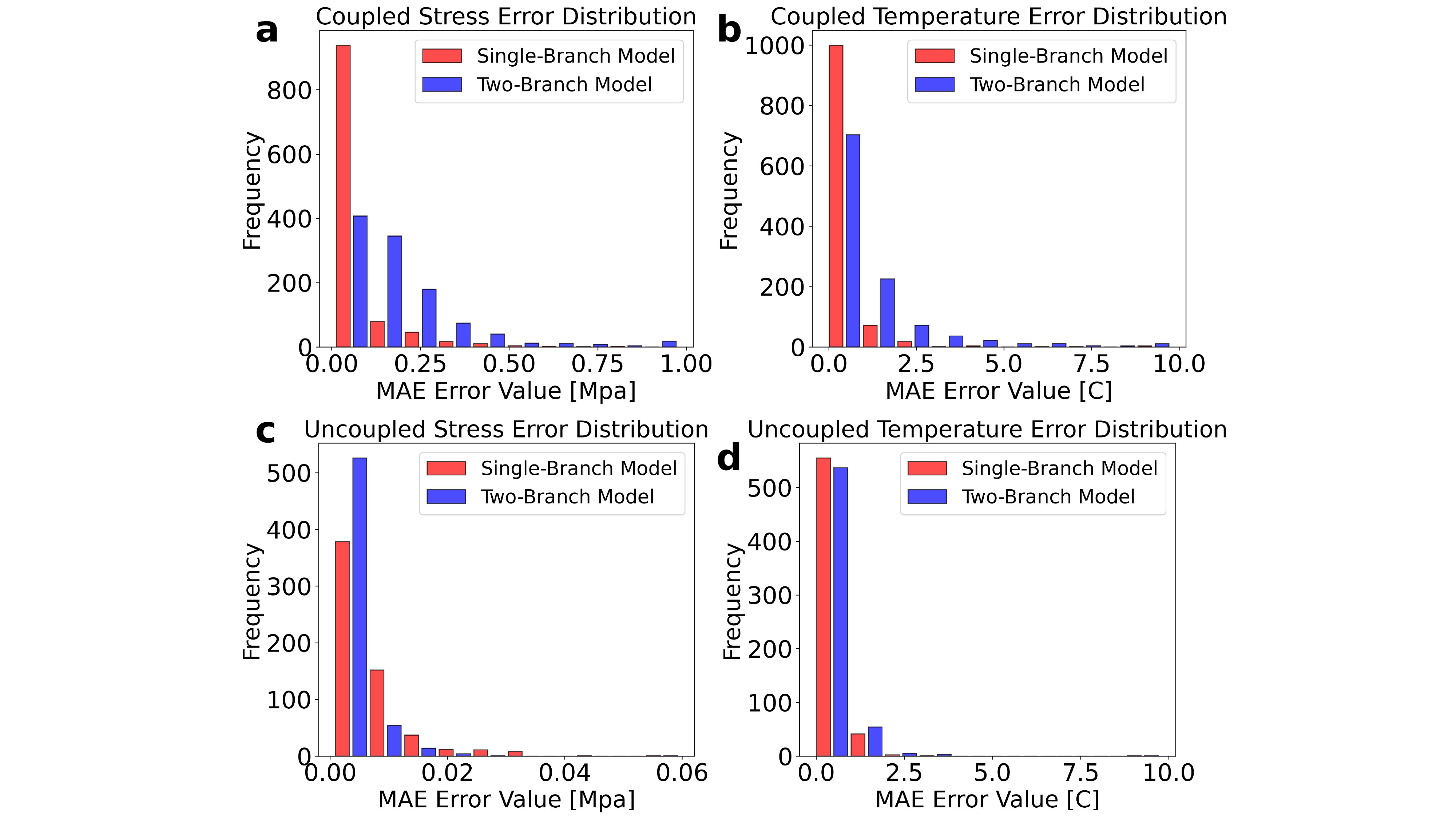}
    \caption{\textbf{Mean absolute error (MAE) distributions for the thermo--mechanical benchmark under coupled and uncoupled regimes.} 
    Coupled setting \textbf{(a), (b)}: histograms of MAE for stress (\textbf{a}) and temperature (\textbf{b}). Single-branch predictions (red) concentrate near zero, whereas two-branch predictions (blue) exhibit long tails, especially for stress. 
    Uncoupled setting \textbf{(c), (d)}: corresponding histograms for stress (\textbf{c}) and temperature (\textbf{d}). With a fixed temperature field, both models are highly accurate, and the two-branch design is slightly favored, consistent with independent inputs benefiting from separate encoders.}
\label{fig:thermo_mech_hist}
\end{figure}

The histogram of test error distributions in Fig.~ \ref{fig:thermo_mech_hist}(a)--(b) further emphasizes the accuracy of single-branch implementation over two-branch implementation, particularly for stress, where two-branch stress predictions (blue columns) have a long tail into larger error areas. 

Similarly to the thermo-electric model, the uncoupled data model with a single-temperature solution passed into the stress model represents a substantially less challenging case for predicting the stress distribution, as stress in this case depends only on the displacement input. The corresponding histogram of test errors in Fig.~\ref{fig:thermo_mech_hist}(c)--(d) shows that the data from the uncoupled dual-physics model slightly favors the two-branch (MIONet) model, which reads and encodes thermal and mechanical inputs separately. 

\begin{table}[htbp]
\caption{Mean absolute error (MAE) for stress and temperature (mean $\pm$ standard deviation over five independent runs). Bold values indicate the lower error between single-branch and two-branch architectures for each physics regime and output field.}
\centering
\begin{tabular}{@{}lcccc@{}}
\toprule
 & \multicolumn{2}{c}{\textbf{Stress (MPa)}} & \multicolumn{2}{c}{\textbf{Temperature ($^\circ$C)}} \\
\cmidrule(lr){2-3} \cmidrule(l){4-5}
\textbf{Physics} & \textbf{Single-Branch} & \textbf{Two-Branch} & \textbf{Single-Branch} & \textbf{Two-Branch} \\
\midrule
Coupled   & $\boldsymbol{0.12 \pm 0.007}$  & $0.219 \pm 0.022$          & $\boldsymbol{0.869 \pm 0.097}$ & $1.466 \pm 0.439$ \\
Uncoupled & $0.008 \pm 0.001$          & $\boldsymbol{0.006 \pm 0.001}$ & $0.703 \pm 0.176$          & $\boldsymbol{0.572 \pm 0.185}$ \\
\bottomrule
\end{tabular}
\label{tab:combined_error_stress_temperature}
\end{table}

\textcolor{black}{Table~\ref{tab:combined_error_stress_temperature} summarizes the thermo-mechanical MAE error results, obtained by repeating training and evaluation five times using five different random seeds while keeping the model architecture, hyperparameters, and training schedules fixed.
The single-branch S-DeepONet, which reads both inputs together and encodes them with shared (coupled) trainable parameters, is more accurate with coupled multiphysics (thermo-mechanical) analysis data. On the other hand, the two-branch (MIONet-based) S-DeepONet model, which reads and encodes each input separately and joins them via Hadamard tensor product, is slightly more accurate for uncoupled physics model data, where each physics depends on its input only.}



\textcolor{black}{
Finally, in addition to the strong empirical evidence demonstrated across different multiphysics examples in operator learning in this work, there is a close analogy between solver coupling in classical numerical analysis and coupling in operator learning. In FEA, a monolithic solver forms a global stiffness matrix where each entry reflects interactions among multiple physical fields. Similarly, in single-branch sequential operator networks, the model jointly processes and encodes inputs from different physics components using shared trainable parameters, thereby learning coupled relationships. Just as monolithic FEA solvers typically yield more accurate results for coupled multiphysics problems than staggered (partitioned) approaches, the single-branch S-DeepONet achieves higher accuracy for coupled multiphysics than the MIONet architecture.}

\subsection*{\textcolor{black}{Robustness to noisy input functions.}}
\textcolor{black}{To assess model deployment-relevant reliability under sensor uncertainty, we perturb the observed input functions with additive Gaussian noise (3--15\%) and propagate $M=100$ noisy realizations per test case through the trained operator.
While errors increase with the applied level of noise, we observe that, in most cases, a single branch performs better on coupled physics, and the two-branch model performs better on uncoupled physics. Specifically, Uncertainty Quantification of model robustness section and Supplementary Note 6 details the Monte Carlo input perturbation protocol, where channel-wise additive Gaussian noise is injected into the observed input histories with noise magnitude scaled by the empirical standard deviation of each channel ($\eta=3$--$15\%$), repeated over $M=100$ realizations. The results showcase the same trend with non-noisy inputs, single branch outperforms its multi-variant and vice versa for the uncoupled case.}


\subsection*{Training Performance and Inference}
The DeepONet architecture, which utilizes fully connected (FNN) networks, requires a relatively large number of trainable parameters to achieve the highest prediction accuracy for the reaction-diffusion case. And even though the multi-branch (MIONet) variant has almost twice the number of trainable parameters, due to highly optimized FNN kernels on the GPU, it did not result in larger training times. 
As listed in Table \ref{tab:5}, for the thermo-mechanical solidification case, S-DeepONet performs recurrent neural network operations in the GRU-based branches and therefore requires a longer training time and a larger number of epochs to achieve high prediction accuracy. However, it was able to complete the entire training in 1.4 hours for a single-branch S-DeepONet and in 2.8 hours for a multi-branch (MIONet) DeepONet. This is particularly encouraging for the single-branch DeepONet, which not only provides more efficient training but also achieves higher accuracy for coupled multiphysics problems.

\textcolor{black}{
Finally, once S-DeepONet is trained, solving a new problem no longer requires repeatedly assembling and solving large sparse systems of nonlinear equations as in FEA. Instead, inference consists of a single forward pass involving a fixed sequence of matrix multiplications and nonlinear activations with pre-trained weights and biases.}

\textcolor{black}{The trained S-DeepONet thermo-mechanical models required 20 and 23 seconds to predict all stress and temperature fields for 1{,}099 test samples, corresponding to prediction times of 0.0185 and 0.020 seconds per unseen sample for the single-branch and two-branch models, respectively. The coupled FEA performed using a highly optimized commercial solver with parallelization on a high-performance computing (HPC) system, required 333~seconds for the same cases. Accordingly, inference with the trained neural networks achieves a speedup of approximately 18{,}000$\times$ compared to the conventional FEA approach.}


\begin{table}[htbp]
\caption{Training Performance.}
\centering
\begin{tabular}{@{}ccccc@{}}
\toprule
\multirow{2}{*}{Model (Architecture)} & \multicolumn{2}{c}{Single-Branch}      & \multicolumn{2}{c}{Multiple-Branch (MIONet)} \\ \cmidrule(l){2-5} 
                                      & Trainable Params & Train Time (Epochs) & Trainable Params    & Train Time (Epochs)    \\ \midrule
ReactDiff (DeepONet)                       & 1.25 M           & 600 sec (110,000)   & 2.5 M               & 658 sec (110,000)      \\
Ther-Mech (S-DeepONet)                     & 0.8 M            & 5,172 sec (300,000) & 1.5 M               & 9,931 sec (300,000)    \\ \bottomrule
\end{tabular}
\label{tab:5}
\end{table}

\section*{Discussion}

\textcolor{black}{Across five independent runs, the results show regime-dependent behavior (Tables~\ref{tab:combined_error_potential_temperature} and~\ref{tab:combined_error_stress_temperature}). In the \textbf{thermo-electric} case, under coupled physics, the single-branch model achieves substantially lower $L_2$ errors for both temperature and electrical potential, approximately 7× and 5.5× lower than the two-branch architecture, respectively. Under uncoupled conditions, both architectures perform comparably, with the two-branch model showing only marginally lower errors. In the \textbf{thermo-mechanical} case, the coupled setting again favors the single-branch model, yielding roughly 45\% lower stress MAE and 40\% lower temperature MAE. Under uncoupled conditions, the two architectures are essentially equivalent. Full numerical tables are provided in the Supplementary Material.}

\textcolor{black}{Within the DeepONet and S-DeepONet architectures studied here, single-branch configurations consistently outperformed two-branch variants on coupled multiphysics tasks. This outcome can be attributed to implicit coupling: concatenating multiple input functions and encoding them in a shared latent space promotes early nonlinear interactions among variables, allowing the network to represent cross-terms intrinsic to coupled PDEs. By contrast, two-branch designs extract features independently and fuse them only at a later stage, which limits information exchange and may underrepresent cross-coupling. This observation parallels the classical distinction between monolithic and staggered solvers in numerical analysis. Monolithic formulations propagate multiphysics interactions through a single global system and generally achieve higher fidelity when couplings are strong. Our empirical findings align with this analogy: single-branch models exhibit superior accuracy in strongly coupled regimes, whereas in uncoupled settings, the performance gap narrows, and dual-branch architectures may remain advantageous due to their modularity and capacity to encode modality-specific priors. We emphasize that this observation is empirical rather than a formal theoretical guarantee, and reflects the behavior of the evaluated operator architectures under the considered benchmarks. We note that these conclusions are demonstrated specifically for DeepONet and S-DeepONet topologies; whether analogous coupling-aware architectural behaviors hold for other operator families, such as Fourier Neural Operators or attention-based architectures \cite{liu2026sequential,park2026sequential}, remains an open question beyond the scope of this study and warrants dedicated cross-family evaluation.  A key limitation of this work is that it focuses exclusively on DeepONet-style operator architectures. While these models are widely used in scientific machine learning, broader validation across alternative operator learning paradigms is required to assess generality.}

\section*{Conclusions}
This study presents a systematic comparison of \emph{single-branch} and \emph{multi-branch} Deep Operator Networks, spanning both classical and sequential (GRU-based S-DeepONet) formulations, across problems of increasing physical complexity, from single-physics to strongly coupled multiphysics regimes. The results yield three central insights:

\begin{enumerate}
    \item \textbf{Architectural design should reflect physical coupling.}  
    For strongly coupled systems, such as thermo-electrical and thermo-mechanical benchmarks, architectures that share learnable parameters across all inputs (\emph{single-branch}) consistently outperform factorized (\emph{multi-branch}) (MIONet-style) designs. This advantage stems from their capacity to model nonlinear inter-field dependencies. In contrast, for uncoupled or single‑physics settings, \emph{multi‑branch} networks provide modest accuracy improvements due to their greater representational flexibility. This observation, within the DeepONet and S-DeepONet families examined here, parallels the classical distinctions between monolithic and partitioned solvers in traditional multiphysics numerical methods, such as finite elements.

    \item \textbf{Operator networks eliminate the runtime bottleneck.}  
    Once trained offline, all evaluated models deliver full-field predictions in milliseconds, achieving up to four orders of magnitude speed-up compared to high-fidelity finite element solvers on modern high-performance computing infrastructure. This acceleration enables applications that were previously infeasible, including real-time digital twins, design-space exploration, and scalable uncertainty propagation.

    \item \textbf{Guidelines for deployment.}  
    The results provide the following architectural observations:  
    (i) shared-parameter branches should be the default for tightly coupled multiphysics problems,  
    (ii) multi-branch encodings remain advantageous in decoupled or single‑physics cases, and  
    (iii) GPU-accelerated training makes operator models practical even for high-resolution, three-dimensional domains \cite{he2024geom}.  
    These observations are scoped to DeepONet and S-DeepONet topologies; their extension to other operator families (e.g., Fourier- or attention-based operators) remains an open direction for future work.
\end{enumerate}

\vspace{1em}

More broadly, this work demonstrates that neural operator architectures, when designed to align with the underlying physics, enable scalable and real-time surrogate models for complex systems that have traditionally been limited by computational constraints. Extending these methods to irregular geometries, incorporating error bounds, and coupling them with optimization and control pipelines represent key directions for future research. Furthermore, sparse sensor settings and corrupted/degraded input scenarios \cite{hossain2024sensor} will be investigated as follow-up work. Together, these advances position operator learning as a promising paradigm for efficient and interpretable modeling in multiphysics simulation and digital twin applications.


\section*{Methods}

\subsection*{Data generation}

\subsubsection*{Reaction–diffusion (Single physics)}
A one-dimensional reaction–diffusion equation characterized by spatially heterogeneous source and reaction coefficients was used as the representative single-physics benchmark. The governing equation is expressed in Eq.~\ref{eq:1}, where $u(x,t)$ denotes the evolving scalar field, $u_0(x)$ is the source term, and $k(x)$ is the spatially varying reaction coefficient. \textcolor{black}{Both $u_0(x)$ and $k(x)$ were independently synthesized as zero-mean Gaussian random fields with prescribed variance and correlation length. For each realization, the variance parameter was randomly drawn from the interval $[1, 50]$, and the covariance kernel followed a Gaussian model with characteristic length scale $l_c = 0.2$. }

The input functions were discretized over a one-dimensional spatial domain $[0,1]$ using 127 second-order finite elements (255 nodes) and evolved over 101 uniform temporal increments using the open-source finite-element package FEniCSx~\cite{baratta2023dolfinx}. The resulting dataset consisted of 10{,}000 input–solution pairs, each corresponding to a unique combination of source and reaction fields used to train and evaluate the operator-learning models.

\subsubsection*{Thermo-electrical (Multi-Physics)}  
Joule heating with coupled electrical conduction and heat diffusion was modeled by incorporating temperature-dependent electrical conductivity and volumetric heat generation induced by the electric field. \textcolor{black}{Two time-dependent drive functions: the external heat input \(Q_{\mathrm{ext}}(t)\) and the charge density \(\rho_e(t)\), were generated as one-dimensional Gaussian random fields, constrained within physically plausible bounds and spatial smoothness with the characteristic correlation length of $0.2$. These inputs were sampled using the same covariance-based approach described in the previous case to ensure spatial correlation and statistical consistency. Random field amplitudes were subsequently nondimensionalized using thermal and electrical property–based scaling factors, namely \(Q_{\mathrm{ext}}^{*} = 3\times10^{5} t_m / (\rho c_p T_s)\) and \(\rho_e^{*} = 10^{7} / (\sigma_0 \phi_s)\), to maintain consistency with the nondimensional form of the coupled governing equations.}

For each input pair $\{Q_{\mathrm{ext}}(t), \rho_e(t)\}$, the coupled system was advanced in FEniCSx~\cite{baratta2023dolfinx} on the same one-dimensional mesh (127 second-order elements, 255 nodes) for 101 uniform time steps, yielding full spatiotemporal fields of temperature $T(x,t)$ and electric potential $\phi(x,t)$. To examine sensitivity to coupling strength, two datasets were generated: a coupled set, where both physical fields evolved concurrently under mutual dependence, and an uncoupled set, where each field was solved independently with the other held fixed or supplied from a single-physics pass. The complete dataset comprised 5{,}000 spatiotemporal samples with paired $(T, \phi)$ fields used for operator training and evaluation.

\subsubsection*{Thermo-mechanical steel solidification (Multi-Physics)} 
Transient thermal evolution with latent heat was treated using an enthalpy formulation, along with quasi-static mechanical equilibrium, under temperature-dependent viscoplastic constitutive behavior representative of steel during continuous casting. The constitutive response was implemented via a UMAT and solved with Abaqus/Standard under generalized plane-strain conditions. The computational domain corresponded to a moving slice along the caster in a Lagrangian frame. Thermal and mechanical boundary conditions were prescribed as time histories that emulate plant operation, with surface heat-flux profiles exhibiting an overall decreasing trend with small fluctuations and displacement histories reflecting mold taper with an overall increasing trend. 

The transient heat conduction with latent heat effects was governed by Equation~\ref{eq:3}, subject to the boundary condition $-k(T)\nabla T = q(t)$ on the cooling surface $\forall x \in \partial\Omega_q$ and initial condition $T(x,0) = T_0$, where $t$ denotes time, $T$ is temperature, $\rho$ is mass density, and $q(t)$ is the time-dependent heat flux. The initial temperature is set to $T_0 = 1540^\circ$C. The functions $H(T)$ and $k(T)$ represent the temperature-dependent specific enthalpy and isotropic thermal conductivity, respectively. The enthalpy $H(T)$ accounts for latent heat released during phase transformations such as solidification and the $\delta$-ferrite–to–austenite transition, introducing strong nonlinearity into the system.

\textcolor{black}{To numerically realize these boundary conditions, the problem domain was discretized into 300 four-node bilinear heat transfer elements (DC2D4) with an element size of 0.1~mm. Equation~\ref{eq:3} was solved for a total of 17~s, corresponding to the residence time of the slice within the mold, using implicit time integration in Abaqus/Standard. The resulting temperature field at the end of the load step was extracted as the ground truth.}


A time-dependent boundary heat flux was generated following the sampling approach~\cite{abueidda2021deep}, in which the flux profile is defined by six control points. The first and last control points correspond to $t = 0$ and $t = 17$~s, respectively, while the four intermediate control-point times $t_{cp}$ were randomly sampled from a uniform distribution over $(0,17)$~s. Based on experimental observations, the heat flux $q_{cp}$ exhibits a decaying temporal trend that can be approximated as
\begin{equation}
q_{cp} = A(t_{cp} + 1)^{-B} + C,
\end{equation}
\textcolor{black}{where $A \in [3, 8]$, $B \in [0.3, 0.7]$, and $C \in [-0.5, 0.5]$ are randomly chosen variables from their respective ranges. The term $C$ acts as an additive random noise component to emulate local fluctuations and nonlinearities commonly observed in the actual heat flux profile, which arise from variable contact and interfacial heat transfer between the mold and the solidifying steel \cite{zappulla2020multiphysics,koric2006efficient}. Once all control-point times and flux values are defined, a radial basis function (RBF) interpolation with a Gaussian kernel is employed to construct a smooth, continuous time-dependent flux profile along the boundary. \cite{he2024sequential}}

\subsection*{\textcolor{black}{Model architectures}}
All models were implemented in DeepXDE (TensorFlow backend) using the DeepONet parameterization introduced in Fig.~\ref{fig:sdon_arch}, with hidden dimension $h = 100$ and zero-initialized biases. Two-branch (MIONet-style) variants apply identical per-input pipelines to each channel and merge their outputs via the Hadamard product, $B = B_d \odot B_m$.

For the reaction–diffusion benchmark, branches are feedforward multilayer perceptron (MLP) since the inputs $u_0(x)$ and $k(x)$ are spatial and not path-dependent; the trunk takes spatiotemporal coordinates $\xi = (x, t)$, and the model produces a single output field ($c=1$). For the electro–thermal and thermo–mechanical benchmarks, branches use a GRU-based encoder–decoder to capture the sequential character of the input histories ($[Q_{\mathrm{ext}}(t), \rho_e(t)]$ and $[q(t), d_n(t)]$, respectively). The trunk takes spatial coordinates $(x, y)$, and each model produces two coupled output fields ($c=2$).

Full layer-by-layer specifications, including MLP widths, GRU unit counts, and trunk dimensions, are provided in the Supplementary Material, and the source code is available on GitHub.

\subsection*{Training and Evaluation protocols}
All models were trained in DeepXDE (TensorFlow backend) with the same optimization settings across all benchmarks to isolate architectural effects: Adam optimizer, initial learning rate $10^{-3}$, inverse–time decay scheduler $(\text{step}=1,\ \text{rate}=10^{-4})$, batch size 64, and a variance–normalized squared–error (COP) loss. The training tasks were executed on an Nvidia A100 GPU within the Delta HPC system \cite{delta_user_doc_2025}.

Iteration budgets differed only by benchmark and coupling: reaction–diffusion (DeepONet) trained for 110{,}000 iterations (single–branch) and 100{,}000 (two–branch); electro–thermal (S–DeepONet) trained for 310{,}000 iterations for both single– and two–branch models in coupled and uncoupled settings; thermo–mechanical solidification (S–DeepONet) trained for 310{,}000 iterations in the coupled case and 110{,}000 in the uncoupled case for both single– and two–branch models.

All results are reported on the held–out 20\% test split; predictions are de–normalized before scoring and visualization. The primary error measures are a per–sample relative \(L_2\) error for full–field comparisons and, where appropriate, the mean absolute error (MAE):
\begin{equation}
L_2 = \frac{\lVert S_{\mathrm{FE}} - S_{\mathrm{pred}} \rVert_2}{\lVert S_{\mathrm{FE}} \rVert_2}, 
\qquad
\mathrm{MAE} = \frac{1}{M}\sum_{j=1}^{M}\bigl|S_{\mathrm{FE}}^{(j)} - S_{\mathrm{pred}}^{(j)}\bigr|,
\end{equation}
where $S_{\mathrm{FE}}$ and $S_{\mathrm{pred}}$ denote the finite–element reference and the model prediction on the same discrete grid, and $M$ is the number of evaluation points per field (spatial or spatiotemporal). For multi–output models $(c=2)$, metrics are computed per field and then summarized across the test set (means and distributions).

In the reaction–diffusion and electro–thermal studies, the primary summary is the relative $L_2$ error over the full field. In the thermo–mechanical solidification study, MAE is employed for stress because frequent near–zero targets in liquid or mushy regions can make relative norms ill–conditioned.

To contextualize averages, representative test instances are visualized at fixed error percentiles. For each benchmark and field, test samples are ranked by per–sample error, and targets, predictions, and absolute–error fields or curves are shown at reaction–diffusion: 55th, 85th, 99th; electro–thermal: best, 55th, 98th; thermo–mechanical: 25th, 45th, 85th. Percentiles are computed independently for each model to reveal characteristic error modes.

Test–set variability is summarized with histograms of per–sample errors for each field. Bin edges and ranges are identical across architectures within a benchmark so that tail behavior is comparable, and the same held–out test split is used for all models.

Discrete norms are evaluated on the native grids of the finite–element outputs: reaction–diffusion errors accumulate over the full $(x,t)$ grid, electro–thermal errors over the two-dimensional spatial mesh, and thermo–mechanical errors along the slice nodes. When multiple time frames are present, the frames shown in figures match those specified for each benchmark.

\subsection*{\textcolor{black}{Uncertainty Quantification of model robustness}}
\label{sec:method_uq}

\textcolor{black}{For each test input function $\mathbf{u}_i(t) \in \mathbb{R}^C$, measurement uncertainty is modeled by generating multiple noisy realizations
\begin{equation}
\tilde{\mathbf{u}}_i^{(m)}(t) = \mathbf{u}_i(t) + \boldsymbol{\epsilon}_i^{(m)}(t),
\qquad m = 1,\dots,M,
\end{equation}
where $\boldsymbol{\epsilon}_i^{(m)}(t)$ denotes stochastic perturbations applied independently to each input channel and time step.}

\textcolor{black}{Specifically, for channel $c \in \{1,\dots,C\}$, the noise is sampled as
\begin{equation}
\epsilon_{i,c}^{(m)}(t) \sim \mathcal{N}\!\left(0,\; \sigma_c^2 \right),
\qquad
\sigma_c = \eta \, \mathrm{std}\!\left(u_{:, :, c}\right),
\end{equation}
where $\eta$ is the prescribed noise level expressed as a fraction of the empirical standard deviation of the corresponding input channel over the test set. This channel-wise scaling ensures that the injected noise is scale-aware and reflects relative sensor uncertainty across heterogeneous input signals.}

\textcolor{black}{For each test case, $M$ independent noisy realizations are generated, representing repeated measurements of the same underlying physical input history. This procedure preserves the original temporal structure while introducing statistically independent measurement errors, enabling Monte Carlo–style uncertainty propagation through the learned operator.}

\textcolor{black}{Each noisy input realization $\tilde{\mathbf{u}}_i^{(m)}(t)$ is passed through the pre-trained neural operator $\mathcal{G}$, yielding
\begin{equation}
\tilde{\mathbf{y}}_i^{(m)}(\mathbf{x})
=
\mathcal{G}\!\left(\tilde{\mathbf{u}}_i^{(m)}(t)\right)(\mathbf{x}),
\qquad m = 1,\dots,M,
\end{equation}
where $\mathbf{x} \in \Omega$ denotes the spatial coordinates at which the solution field is evaluated. The operator $\mathcal{G} : \mathcal{U} \rightarrow \mathcal{Y}$ maps time-dependent input functions to spatially distributed physical response fields.}

\textcolor{black}{Collecting the outputs across all Monte Carlo realizations yields a tensor of shape
\[
[M,\, N,\, ch],
\]
where $M$ is the number of noisy input realizations, $N$ is the number of spatial discretization points, and $ch$ is the number of physical output quantities. For the thermo-mechanical benchmark, the operator predicts the temperature and stress fields on a fixed spatial grid with $N=602$ nodes, resulting in an output tensor of size $[M, 602, 2]$. In all experiments, $M$ is set to $100$. This procedure is repeated for all test input functions $\{\mathbf{u}_i(t)\}_{i=1}^{N_{\mathrm{test}}}$.}

\textcolor{black}{For each noisy realization $\tilde{\mathbf{u}}_i^{(m)}(t)$, the prediction error is quantified via the mean absolute error (MAE) between the operator output and the ground-truth solution:
\begin{equation}
e_i^{(m)} = \frac{1}{N} \sum_{j=1}^{N}
\left|
\mathcal{G}\!\left(\tilde{\mathbf{u}}_i^{(m)}(t)\right)(\mathbf{x}_j)
-
\mathbf{y}_i^{\mathrm{gt}}(\mathbf{x}_j)
\right|.
\end{equation}
This yields, for each test case $i$, an ensemble of $M$ MAE values $\{e_i^{(1)}, \dots, e_i^{(M)}\}$ that characterizes the sensitivity of the prediction accuracy to input measurement noise.}

\textcolor{black}{To summarize the error distribution across test cases and Monte Carlo realizations, we report the sample mean of the MAE as a measure of central tendency, together with the interquartile range (IQR), defined by the 25th and 75th percentiles, to capture the spread of the error distribution without sensitivity to outliers. These statistics are computed separately for each output channel (stress and temperature). The results are presented in the Supplementary Material.}

\section*{Data Availability Statement}
All datasets used in this study are publicly available through the Hugging Face dataset repository at \url{https://huggingface.co/datasets/jaewan-wod33/Single_vs_Multiple_Branches_in_S_DeepONet}.

\section*{Code Availability Statement}
All code supporting the findings of this study are available through the GitHub repository at \url{https://github.com/benjamin0303/Single_vs_Multiple_Branches_in_S_DeepONet}, ensuring transparency, reproducibility, and long-term accessibility.

\bibliographystyle{unsrt}  
\bibliography{references}

\begin{thebibliography}{10}

\bibitem{kobayashi2024ai}
Kazuma Kobayashi et~al.
\newblock Ai-driven non-intrusive uncertainty quantification of advanced
  nuclear fuels for digital twin-enabling technology.
\newblock {\em Progress in Nuclear Energy}, 172:105177, 2024.

\bibitem{kumar2022multi}
Dinesh Kumar et~al.
\newblock Multi-criteria decision making under uncertainties in composite
  materials selection and design.
\newblock {\em Composite Structures}, 279:114680, 2022.

\bibitem{pagani2021enabling}
Stefano Pagani and Andrea Manzoni.
\newblock Enabling forward uncertainty quantification and sensitivity analysis
  in cardiac electrophysiology by reduced order modeling and machine learning.
\newblock {\em International Journal for Numerical Methods in Biomedical
  Engineering}, 37(6):e3450, 2021.

\bibitem{shahane2022surrogate}
Shantanu Shahane, Erman Guleryuz, Diab~W Abueidda, Allen Lee, Joe Liu, Xin Yu,
  Raymond Chiu, Seid Koric, Narayana~R Aluru, and Placid~M Ferreira.
\newblock Surrogate neural network model for sensitivity analysis and
  uncertainty quantification of the mechanical behavior in the optical
  lens-barrel assembly.
\newblock {\em Computers \& Structures}, 270:106843, 2022.

\bibitem{liu2024adaptive}
Qibang Liu, Diab Abueidda, Sagar Vyas, Yuan Gao, Seid Koric, and Philippe~H
  Geubelle.
\newblock Adaptive data-driven deep-learning surrogate model for frontal
  polymerization in dicyclopentadiene.
\newblock {\em The Journal of Physical Chemistry B}, 128(5):1220--1230, 2024.

\bibitem{kushwaha2024advanced}
Shashank Kushwaha, Jaewan Park, Seid Koric, Junyan He, Iwona Jasiuk, and Diab
  Abueidda.
\newblock Advanced deep operator networks to predict multiphysics solution
  fields in materials processing and additive manufacturing.
\newblock {\em Additive Manufacturing}, 88:104266, 2024.

\bibitem{goli2020chemnet}
Elyas Goli, Sagar Vyas, Seid Koric, Nahil Sobh, and Philippe~H Geubelle.
\newblock Chemnet: A deep neural network for advanced composites manufacturing.
\newblock {\em The Journal of Physical Chemistry B}, 124(42):9428--9437, 2020.

\bibitem{kollmann2020deep}
Hunter~T Kollmann, Diab~W Abueidda, Seid Koric, Erman Guleryuz, and Nahil~A
  Sobh.
\newblock Deep learning for topology optimization of 2d metamaterials.
\newblock {\em Materials \& Design}, 196:109098, 2020.

\bibitem{bastek2023inverse}
Jan-Hendrik Bastek and Dennis~M Kochmann.
\newblock Inverse design of nonlinear mechanical metamaterials via video
  denoising diffusion models.
\newblock {\em Nature Machine Intelligence}, 5(12):1466--1475, 2023.

\bibitem{raissi2019physics}
Maziar Raissi, Paris Perdikaris, and George~E Karniadakis.
\newblock Physics-informed neural networks: A deep learning framework for
  solving forward and inverse problems involving nonlinear partial differential
  equations.
\newblock {\em Journal of Computational physics}, 378:686--707, 2019.

\bibitem{sun2020surrogate}
Luning Sun, Han Gao, Shaowu Pan, and Jian-Xun Wang.
\newblock Surrogate modeling for fluid flows based on physics-constrained deep
  learning without simulation data.
\newblock {\em Computer Methods in Applied Mechanics and Engineering},
  361:112732, 2020.

\bibitem{he2023deep}
Junyan He, Diab Abueidda, Rashid~Abu Al-Rub, Seid Koric, and Iwona Jasiuk.
\newblock A deep learning energy-based method for classical elastoplasticity.
\newblock {\em International Journal of Plasticity}, 162:103531, 2023.

\bibitem{fuhg2022mixed}
Jan~N Fuhg and Nikolaos Bouklas.
\newblock The mixed deep energy method for resolving concentration features in
  finite strain hyperelasticity.
\newblock {\em Journal of Computational Physics}, 451:110839, 2022.

\bibitem{nguyen2020deep}
Vien~Minh Nguyen-Thanh, Xiaoying Zhuang, and Timon Rabczuk.
\newblock A deep energy method for finite deformation hyperelasticity.
\newblock {\em European Journal of Mechanics-A/Solids}, 80:103874, 2020.

\bibitem{krishnapriyan2021characterizing}
Aditi Krishnapriyan, Amir Gholami, Shandian Zhe, Robert Kirby, and Michael~W
  Mahoney.
\newblock Characterizing possible failure modes in physics-informed neural
  networks.
\newblock {\em Advances in neural information processing systems},
  34:26548--26560, 2021.

\bibitem{wang2022and}
Sifan Wang, Xinling Yu, and Paris Perdikaris.
\newblock When and why pinns fail to train: A neural tangent kernel
  perspective.
\newblock {\em Journal of Computational Physics}, 449:110768, 2022.

\bibitem{kovachki2023neural}
Nikola Kovachki, Zongyi Li, Burigede Liu, Kamyar Azizzadenesheli, Kaushik
  Bhattacharya, Andrew Stuart, and Anima Anandkumar.
\newblock Neural operator: Learning maps between function spaces with
  applications to pdes.
\newblock {\em Journal of Machine Learning Research}, 24(89):1--97, 2023.

\bibitem{azizzadenesheli2024neural}
Kamyar Azizzadenesheli, Nikola Kovachki, Zongyi Li, Miguel Liu-Schiaffini, Jean
  Kossaifi, and Anima Anandkumar.
\newblock Neural operators for accelerating scientific simulations and design.
\newblock {\em Nature Reviews Physics}, 6(5):320--328, 2024.

\bibitem{li2020fourier}
Zongyi Li, Nikola~Borislavov Kovachki, Kamyar Azizzadenesheli, Kaushik
  Bhattacharya, Andrew Stuart, Anima Anandkumar, et~al.
\newblock Fourier neural operator for parametric partial differential
  equations.
\newblock In {\em International Conference on Learning Representations}, 2021.

\bibitem{liu2025enhancing}
Chaoyu Liu, Davide Murari, Chris Budd, Lihao Liu, and Carola-Bibiane
  Sch{\"o}nlieb.
\newblock Enhancing fourier neural operators with local spatial features.
\newblock {\em arXiv preprint arXiv:2503.17797}, 2025.

\bibitem{xiao2024amortized}
Zipeng Xiao, Siqi Kou, Hao Zhongkai, Bokai Lin, and Zhijie Deng.
\newblock Amortized fourier neural operators.
\newblock {\em Advances in Neural Information Processing Systems},
  37:115001--115020, 2024.

\bibitem{tripura2022wavelet}
Tapas Tripura and Souvik Chakraborty.
\newblock Wavelet neural operator for solving parametric partial differential
  equations in computational mechanics problems.
\newblock {\em Computer Methods in Applied Mechanics and Engineering},
  404:115783, 2023.

\bibitem{liu2024mitigating}
Xinliang Liu, Bo~Xu, Shuhao Cao, and Lei Zhang.
\newblock Mitigating spectral bias for the multiscale operator learning.
\newblock {\em Journal of Computational Physics}, 506:112944, 2024.

\bibitem{qin2024toward}
Shaoxiang Qin, Fuyuan Lyu, Wenhui Peng, Dingyang Geng, Ju~Wang, Naiping Gao,
  Xue Liu, and Liangzhu~Leon Wang.
\newblock Toward a better understanding of fourier neural operators: Analysis
  and improvement from a spectral perspective.
\newblock {\em arXiv e-prints}, pages arXiv--2404, 2024.

\bibitem{wen2022u}
Gege Wen, Zongyi Li, Kamyar Azizzadenesheli, Anima Anandkumar, and Sally~M
  Benson.
\newblock U-fno—an enhanced fourier neural operator-based deep-learning model
  for multiphase flow.
\newblock {\em Advances in Water Resources}, 163:104180, 2022.

\bibitem{tranfactorized}
Alasdair Tran, Alexander Mathews, Lexing Xie, and Cheng~Soon Ong.
\newblock Factorized fourier neural operators.
\newblock In {\em The Eleventh International Conference on Learning
  Representations}, 2023.

\bibitem{zhao2023enhancing}
Xuanle Zhao, Yue Sun, Tielin Zhang, and Bo~Xu.
\newblock Enhancing solutions for complex pdes: Introducing complementary
  convolution and equivariant attention in fourier neural operators.
\newblock {\em arXiv preprint arXiv:2311.12902}, 2023.

\bibitem{liu2022ht}
Xinliang Liu, Bo~Xu, and Lei Zhang.
\newblock Ht-net: Hierarchical transformer based operator learning model for
  multiscale pdes.
\newblock 2022.

\bibitem{zhang2024d2no}
Zecheng Zhang, Christian Moya, Lu~Lu, Guang Lin, and Hayden Schaeffer.
\newblock D2no: Efficient handling of heterogeneous input function spaces with
  distributed deep neural operators.
\newblock {\em Computer Methods in Applied Mechanics and Engineering},
  428:117084, 2024.

\bibitem{hao2023gnot}
Zhongkai Hao, Zhengyi Wang, Hang Su, Chengyang Ying, Yinpeng Dong, Songming
  Liu, Ze~Cheng, Jian Song, and Jun Zhu.
\newblock Gnot: A general neural operator transformer for operator learning.
\newblock In {\em International Conference on Machine Learning}, pages
  12556--12569. PMLR, 2023.

\bibitem{ovadia2024vito}
Oded Ovadia, Adar Kahana, Panos Stinis, Eli Turkel, Dan Givoli, and George~Em
  Karniadakis.
\newblock Vito: Vision transformer-operator.
\newblock {\em Computer Methods in Applied Mechanics and Engineering},
  428:117109, 2024.

\bibitem{lu2021learning}
Lu~Lu, Pengzhan Jin, Guofei Pang, Zhongqiang Zhang, and George~Em Karniadakis.
\newblock Learning nonlinear operators via deeponet based on the universal
  approximation theorem of operators.
\newblock {\em Nature machine intelligence}, 3(3):218--229, 2021.

\bibitem{chen1995universal}
Tianping Chen and Hong Chen.
\newblock Universal approximation to nonlinear operators by neural networks
  with arbitrary activation functions and its application to dynamical systems.
\newblock {\em IEEE transactions on neural networks}, 6(4):911--917, 1995.

\bibitem{wang2021learning}
Sifan Wang, Hanwen Wang, and Paris Perdikaris.
\newblock Learning the solution operator of parametric partial differential
  equations with physics-informed deeponets.
\newblock {\em Science advances}, 7(40):eabi8605, 2021.

\bibitem{mandl2025separable}
Luis Mandl, Somdatta Goswami, Lena Lambers, and Tim Ricken.
\newblock Separable physics-informed deeponet: Breaking the curse of
  dimensionality in physics-informed machine learning.
\newblock {\em Computer Methods in Applied Mechanics and Engineering},
  434:117586, 2025.

\bibitem{he2024sequential}
Junyan He, Shashank Kushwaha, Jaewan Park, Seid Koric, Diab Abueidda, and Iwona
  Jasiuk.
\newblock Sequential deep operator networks (s-deeponet) for predicting
  full-field solutions under time-dependent loads.
\newblock {\em Engineering Applications of Artificial Intelligence},
  127:107258, 2024.

\bibitem{koric2024deep}
Seid Koric, Asha Viswantah, Diab~W Abueidda, Nahil~A Sobh, and Kamran Khan.
\newblock Deep learning operator network for plastic deformation with variable
  loads and material properties.
\newblock {\em Engineering with Computers}, 40(2):917--929, 2024.

\bibitem{goswami2022physics}
Somdatta Goswami, Minglang Yin, Yue Yu, and George~Em Karniadakis.
\newblock A physics-informed variational deeponet for predicting crack path in
  quasi-brittle materials.
\newblock {\em Computer Methods in Applied Mechanics and Engineering},
  391:114587, 2022.

\bibitem{zhao2023learning}
Tun Zhao, Weiqi Qian, Jie Lin, Hai Chen, Houjun Ao, Gong Chen, and Lei He.
\newblock Learning mappings from iced airfoils to aerodynamic coefficients
  using a deep operator network.
\newblock {\em Journal of Aerospace Engineering}, 36(5):04023035, 2023.

\bibitem{xu2023training}
Liang Xu, Haigang Zhang, and Minghui Zhang.
\newblock Training a deep operator network as a surrogate solver for
  two-dimensional parabolic-equation models.
\newblock {\em The Journal of the Acoustical Society of America},
  154(5):3276--3284, 2023.

\bibitem{haghighat2024deeponet}
Ehsan Haghighat, Umair bin Waheed, and George Karniadakis.
\newblock En-deeponet: An enrichment approach for enhancing the expressivity of
  neural operators with applications to seismology.
\newblock {\em Computer Methods in Applied Mechanics and Engineering},
  420:116681, 2024.

\bibitem{kobayashi2024improved}
Kazuma Kobayashi, James Daniell, and Syed~Bahauddin Alam.
\newblock Improved generalization with deep neural operators for engineering
  systems: Path towards digital twin.
\newblock {\em Engineering Applications of Artificial Intelligence},
  131:107844, 2024.

\bibitem{kobayashi2025proxies}
Kazuma Kobayashi, Samrendra Roy, Seid Koric, Diab Abueidda, and Syed~Bahauddin
  Alam.
\newblock From proxies to fields: Spatiotemporal reconstruction of global
  radiation from sparse sensor sequences.
\newblock {\em arXiv preprint arXiv:2506.12045}, 2025.

\bibitem{kobayashi2024deep}
Kazuma Kobayashi et~al.
\newblock Deep neural operator-driven real-time inference to enable digital
  twin solutions for nuclear energy systems.
\newblock {\em Nature Scientific reports}, 14(1):2101, 2024.

\bibitem{hossain2025virtual}
Raisa Hossain, Farid Ahmed, Kazuma Kobayashi, Seid Koric, Diab Abueidda, and
  Syed~Bahauddin Alam.
\newblock Virtual sensing-enabled digital twin framework for real-time
  monitoring of nuclear systems leveraging deep neural operators.
\newblock {\em npj Materials Degradation}, 9(1):21, 2025.

\bibitem{sahin2024deep}
Izzet Sahin, Christian Moya, Amirhossein Mollaali, Guang Lin, and Guillermo
  Paniagua.
\newblock Deep operator learning-based surrogate models with uncertainty
  quantification for optimizing internal cooling channel rib profiles.
\newblock {\em International Journal of Heat and Mass Transfer}, 219:124813,
  2024.

\bibitem{koric2023data}
Seid Koric and Diab~W Abueidda.
\newblock Data-driven and physics-informed deep learning operators for solution
  of heat conduction equation with parametric heat source.
\newblock {\em International Journal of Heat and Mass Transfer}, 203:123809,
  2023.

\bibitem{lee2024training}
Sanghyun Lee and Yeonjong Shin.
\newblock On the training and generalization of deep operator networks.
\newblock {\em SIAM Journal on Scientific Computing}, 46(4):C273--C296, 2024.

\bibitem{howard2023multifidelity}
Amanda~A Howard, Mauro Perego, George~Em Karniadakis, and Panos Stinis.
\newblock Multifidelity deep operator networks for data-driven and
  physics-informed problems.
\newblock {\em Journal of Computational Physics}, 493:112462, 2023.

\bibitem{cai2021deepm}
Shengze Cai, Zhicheng Wang, Lu~Lu, Tamer~A Zaki, and George~Em Karniadakis.
\newblock Deepm\&mnet: Inferring the electroconvection multiphysics fields
  based on operator approximation by neural networks.
\newblock {\em Journal of Computational Physics}, 436:110296, 2021.

\bibitem{mao2021deepm}
Zhiping Mao, Lu~Lu, Olaf Marxen, Tamer~A Zaki, and George~Em Karniadakis.
\newblock Deepm\&mnet for hypersonics: Predicting the coupled flow and
  finite-rate chemistry behind a normal shock using neural-network
  approximation of operators.
\newblock {\em Journal of computational physics}, 447:110698, 2021.

\bibitem{jiang2024fourier}
Zhongyi Jiang, Min Zhu, and Lu~Lu.
\newblock Fourier-mionet: Fourier-enhanced multiple-input neural operators for
  multiphase modeling of geological carbon sequestration.
\newblock {\em Reliability Engineering \& System Safety}, 251:110392, 2024.

\bibitem{li2024physics}
Zongyi Li, Hongkai Zheng, Nikola Kovachki, David Jin, Haoxuan Chen, Burigede
  Liu, Kamyar Azizzadenesheli, and Anima Anandkumar.
\newblock Physics-informed neural operator for learning partial differential
  equations.
\newblock {\em ACM/JMS Journal of Data Science}, 1(3):1--27, 2024.

\bibitem{yuan2025high}
Biao Yuan, He~Wang, Yanjie Song, Ana Heitor, and Xiaohui Chen.
\newblock High-fidelity multiphysics modelling for rapid predictions using
  physics-informed parallel neural operator.
\newblock {\em arXiv preprint arXiv:2502.19543}, 2025.

\bibitem{rahman2024pretraining}
Md~Ashiqur Rahman, Robert~Joseph George, Mogab Elleithy, Daniel Leibovici,
  Zongyi Li, Boris Bonev, Colin White, Julius Berner, Raymond~A Yeh, Jean
  Kossaifi, et~al.
\newblock Pretraining codomain attention neural operators for solving
  multiphysics pdes.
\newblock {\em Advances in Neural Information Processing Systems},
  37:104035--104064, 2024.

\bibitem{zhao2025diffeomorphism}
Zhiwei Zhao, Changqing Liu, Yingguang Li, Zhibin Chen, and Xu~Liu.
\newblock Diffeomorphism neural operator for various domains and parameters of
  partial differential equations.
\newblock {\em Communications Physics}, 8(1):15, 2025.

\bibitem{liu2024deep}
Ning Liu, Xuxiao Li, Manoj~R Rajanna, Edward~W Reutzel, Brady Sawyer, Prahalada
  Rao, Jim Lua, Nam Phan, and Yue Yu.
\newblock Deep neural operator enabled digital twin modeling for additive
  manufacturing.
\newblock {\em Advances in Computational Science and Engineering},
  2(3):174--201, 2024.

\bibitem{kobayashi2024explainable}
Kazuma Kobayashi et~al.
\newblock Explainable, interpretable, and trustworthy {AI} for an intelligent
  digital twin: A case study on remaining useful life.
\newblock {\em Engineering Applications of Artificial Intelligence},
  129:107620, 2024.

\bibitem{daniell2025digital}
James Daniell et~al.
\newblock Digital twin-centered hybrid data-driven multi-stage deep learning
  framework for enhanced nuclear reactor power prediction.
\newblock {\em Energy and AI}, 19:100450, 2025.

\bibitem{roy2026adversarial}
Samrendra Roy et~al.
\newblock Adversarial vulnerabilities in neural operator digital twins:
  Gradient-free attacks on nuclear thermal-hydraulic surrogates.
\newblock {\em arXiv preprint arXiv:2603.22525}, 2026.

\bibitem{howes2026graph}
William Howes et~al.
\newblock Graph neural operator towards edge deployability and portability for
  sparse-to-dense, real-time virtual sensing on irregular grids.
\newblock {\em arXiv preprint arXiv:2604.01802}, 2026.

\bibitem{yoo2025cross}
Jay~Phil Yoo et~al.
\newblock Sensing without colocation: Operator-based virtual instrumentation
  for domains beyond physical reach.
\newblock {\em arXiv preprint arXiv:2510.18041}, 2025.

\bibitem{alam2019neutronic}
S.~Alam et~al.
\newblock Neutronic investigation of alternative \& composite burnable poisons
  for the soluble-boron-free and long life civil marine small modular reactor
  cores.
\newblock {\em Scientific reports}, 9(1):19591, 2019.

\bibitem{alam2019small1}
S.~Alam et~al.
\newblock Small modular reactor core design for civil marine propulsion using
  micro-heterogeneous duplex fuel. part i: Assembly-level analysis.
\newblock {\em Nuclear Engineering and Design}, 346:157--175, 2019.

\bibitem{alam2019small2}
S.~Alam et~al.
\newblock Small modular reactor core design for civil marine propulsion using
  micro-heterogeneous duplex fuel. part ii: Whole-core analysis.
\newblock {\em Nuclear Engineering and Design}, 346:176--191, 2019.

\bibitem{ahmed2021numerical}
Farid Ahmed et~al.
\newblock Numerical investigation of the thermo-hydraulic performance of
  water-based nanofluids in a dimpled channel flow using al2o3, cuo, and hybrid
  al2o3--cuo as nanoparticles.
\newblock {\em Heat transfer}, 50(5):5080--5105, 2021.

\bibitem{yang2025multiphysics}
Changfan Yang, Lichen Bai, Yinpeng Wang, Shufei Zhang, and Zeke Xie.
\newblock Multiphysics bench: Benchmarking and investigating scientific machine
  learning for multiphysics pdes.
\newblock {\em arXiv preprint arXiv:2505.17575}, 2025.

\bibitem{jin2022mionet}
Pengzhan Jin, Shuai Meng, and Lu~Lu.
\newblock Mionet: Learning multiple-input operators via tensor product.
\newblock {\em SIAM Journal on Scientific Computing}, 44(6):A3490--A3514, 2022.

\bibitem{muller2022gstools}
Sebastian M{\"u}ller, Lennart Sch{\"u}ler, Alraune Zech, and Falk He{\ss}e.
\newblock Gstools v1. 3: a toolbox for geostatistical modelling in python.
\newblock {\em Geoscientific Model Development}, 15(7):3161--3182, 2022.

\bibitem{baratta2023dolfinx}
Igor~A Baratta, Joseph~P Dean, J{\o}rgen~S Dokken, Michal Habera, Jack HALE,
  Chris~N Richardson, Marie~E Rognes, Matthew~W Scroggs, Nathan Sime, and
  Garth~N Wells.
\newblock Dolfinx: the next generation fenics problem solving environment.
\newblock {\em Preprint}, 2023.

\bibitem{Antoulinakis2016}
Foivos Antoulinakis, Daniel Chernin, Peng Zhang, and Y.~Y. Lau.
\newblock Effects of temperature dependence of electrical and thermal
  conductivities on the joule heating of a one dimensional conductor.
\newblock {\em Journal of Applied Physics}, 120(13):135105, 2016.

\bibitem{Kalungi2025}
Paul Kalungi and James Menart.
\newblock Electrochemical--thermal model of a lithium-ion battery.
\newblock {\em Energies}, 18(7):1764, 2025.

\bibitem{LopezMolina2016}
J.~A.~Lopez Molina, M.~J. Rivera, and Enrique~J. Berjano.
\newblock Electrical-thermal analytical modeling of monopolar rf thermal
  ablation of biological tissues: determining the circumstances under which
  tissue temperature reaches a steady state.
\newblock {\em Mathematical Biosciences and Engineering}, 13(2):281--301, 2016.

\bibitem{Sun2018}
Ze~Sun, Liwei Cai, Haiou Ni, Gui-Min Lu, and Jian-Guo Yu.
\newblock Coupled electro-thermal field in a high current electrolysis cell or
  liquid metal batteries.
\newblock {\em Royal Society Open Science}, 5(2):171309, 2018.

\bibitem{Feulvarch2004}
E.~Feulvarch, V.~Robin, and J.-M. Bergheau.
\newblock Resistance spot welding simulation: a general finite element
  formulation of electrothermal contact conditions.
\newblock {\em Journal of Materials Processing Technology}, 153--154:436--441,
  2004.

\bibitem{thomas2018review}
Brian~G Thomas.
\newblock Review on modeling and simulation of continuous casting.
\newblock {\em steel research international}, 89(1):1700312, 2018.

\bibitem{Louhenkilpi2014}
Seppo Louhenkilpi.
\newblock Continuous casting of steel.
\newblock In Seshadri Seetharaman, editor, {\em Treatise on Process
  Metallurgy}, pages 373--434. Elsevier, Boston, 2014.

\bibitem{koric2010multiphysics}
Seid Koric, Lance~C Hibbeler, Rui Liu, and Brian~G Thomas.
\newblock Multiphysics model of metal solidification on the continuum level.
\newblock {\em Numerical Heat Transfer, Part B: Fundamentals}, 58(6):371--392,
  2010.

\bibitem{kozlowski1992simple}
Patrick~F Kozlowski, Brian~G Thomas, Jean~A Azzi, and Hao Wang.
\newblock Simple constitutive equations for steel at high temperature.
\newblock {\em Metallurgical Transactions A}, 23:903--918, 1992.

\bibitem{zhu1996coupled}
Hong Zhu.
\newblock Coupled thermo-mechanical finite-element model with application to
  initial solidification.
\newblock {\em The University of Illinois at Urbana-Champaign}, 1996.

\bibitem{koric2006efficient}
Seid Koric and Brian~G Thomas.
\newblock Efficient thermo-mechanical model for solidification processes.
\newblock {\em International journal for numerical methods in engineering},
  66(12):1955--1989, 2006.

\bibitem{abaqus2022}
{Dassault Syst\`emes}.
\newblock {\em Abaqus/Standard User's Manual, Version 2022}.
\newblock Dassault Syst\`emes Simulia Corp., Providence, RI, 2022.
\newblock Accessed via licensed software documentation.

\bibitem{zappulla2020multiphysics}
Matthew~LS Zappulla, Seong-Mook Cho, Seid Koric, Hyoung-Jun Lee, Seon-Hyo Kim,
  and Brian~G Thomas.
\newblock Multiphysics modeling of continuous casting of stainless steel.
\newblock {\em Journal of Materials Processing Technology}, 278:116469, 2020.

\bibitem{liu2026sequential}
Qibang Liu, Weiheng Zhong, Seid Koric, and Hadi Meidani.
\newblock Sequential neural operator transformer for high-fidelity surrogates
  of time-dependent non-linear partial differential equations.
\newblock {\em Engineering Applications of Artificial Intelligence},
  172:114428, 2026.

\bibitem{park2026sequential}
Jaewan Park, Kazuma Kobayashi, Qibang Liu, Amar~Alem Koric, Diab~W Abueidda,
  Syed~Bahauddin Alam, and Seid Koric.
\newblock Sequential deep operator neural networks for plastic and
  thermo-viscoplastic transient material behavior.
\newblock {\em International Journal of Plasticity}, 201:104689, 2026.

\bibitem{he2024geom}
Junyan He, Seid Koric, Diab Abueidda, Ali Najafi, and Iwona Jasiuk.
\newblock Geom-deeponet: A point-cloud-based deep operator network for field
  predictions on 3d parameterized geometries.
\newblock {\em Computer Methods in Applied Mechanics and Engineering},
  429:117130, 2024.

\bibitem{hossain2024sensor}
R~Hossain et~al.
\newblock Sensor degradation in nuclear reactor pressure vessels: the
  overlooked factor in remaining useful life prediction.
\newblock {\em npj Materials Degradation}, 8(1):71, 2024.

\bibitem{abueidda2021deep}
Diab~W Abueidda, Seid Koric, Nahil~A Sobh, and Huseyin Sehitoglu.
\newblock Deep learning for plasticity and thermo-viscoplasticity.
\newblock {\em International Journal of Plasticity}, 136:102852, 2021.

\bibitem{delta_user_doc_2025}
{National Center for Supercomputing Applications (NCSA)}.
\newblock {Delta} user documentation.
\newblock
  \url{https://docs.ncsa.illinois.edu/systems/delta/en/latest/index.html},
  2025.
\newblock Accessed: January 25, 2025.

\end{thebibliography}





\section*{Acknowledgments}
The authors thank the National Center for Supercomputing Applications (NCSA), particularly the Research Consulting Directorate, and the Center for Artificial Intelligence Innovation (CAII) at the University of Illinois Urbana-Champaign for their support. Additional computational support was provided by the Illinois Computes project, a joint initiative of the University of Illinois Urbana-Champaign and the University of Illinois System.

\section*{Funding Statement}
This research used the Delta and DeltaAI advanced computing and data resources, supported by the National Science Foundation (NSF) under awards OAC-2005572 and OAC-2320345, and by the State of Illinois. Delta and DeltaAI are joint efforts of the University of Illinois Urbana-Champaign and the National Center for Supercomputing Applications (NCSA). This work was supported in part by the NSF under grant 209875. S. B. Alam acknowledges support from the U.S. Nuclear Regulatory Commission under grant numbers 31310025M0012 and 31310024M0041.

\section*{Author Information}
\textbf{Affiliations} \\
\textbf{Jaewan Park}: National Center for Supercomputing Applications, University of Illinois at Urbana-Champaign, Urbana, IL, USA; The Grainger College of Engineering, Mechanical Science and Engineering, University of Illinois at Urbana-Champaign, Urbana, IL, USA. \\
\textbf{Kazuma Kobayashi}: The Grainger College of Engineering, Nuclear, Plasma \& Radiological Engineering, University of Illinois at Urbana-Champaign, Urbana, IL, USA. \\
\textbf{Qibang Liu}: Department of Industrial and Manufacturing Systems Engineering, Kansas State University, Manhattan, KS, USA. \\
\textbf{Diab Abueidda}: Civil and Urban Engineering Department, New York University Abu Dhabi, UAE. \\
\textbf{Seid Koric}: National Center for Supercomputing Applications, University of Illinois at Urbana-Champaign, Urbana, IL, USA; The Grainger College of Engineering, Mechanical Science and Engineering, University of Illinois at Urbana-Champaign, Urbana, IL, USA. \\
\textbf{Syed Bahauddin Alam}: National Center for Supercomputing Applications, University of Illinois at Urbana-Champaign, Urbana, IL, USA; The Grainger College of Engineering, Nuclear, Plasma \& Radiological Engineering, University of Illinois at Urbana-Champaign, Urbana, IL, USA.

\textbf{Author Contributions}\\
J.P. and K.K. conceptualized the study, developed the methodology, and performed data analysis. Q.L. and D.A. developed the methodology and performed data analysis. S.K. and S.B.A. provided conceptualization, resources, acquired funding, and supervised the research. All authors contributed to the writing and editing of the manuscript.

\textbf{Corresponding Author} \\
Correspondence to Syed Bahauddin Alam (alams@illinois.edu).

\section*{Ethics Declarations}
\textbf{Disclosure of Interests.}\\
The authors have no competing interests to declare that are relevant to the content of this article. 

\section*{Additional information}
Supplementary technical details, extended results, and training logs are provided in the Supplementary Information.

\clearpage
\setcounter{figure}{0}
\setcounter{table}{0}
\renewcommand{\thefigure}{S\arabic{figure}}
\renewcommand{\thetable}{S\arabic{table}}
\renewcommand{\figurename}{Supplementary Figure}
\renewcommand{\tablename}{Supplementary Table}

\section*{Supplementary Information}

This supplementary document includes additional technical content and analysis not covered in the main manuscript. It contains detailed training curves, extended results, and supporting materials that enhance reproducibility and interpretation of the study.

\section*{Supplementary Note 1: Deep Operator Network}
\label{sup:deeponet}

\begin{figure}[!htbp]
    \centering
    \includegraphics[width=0.7\textwidth]{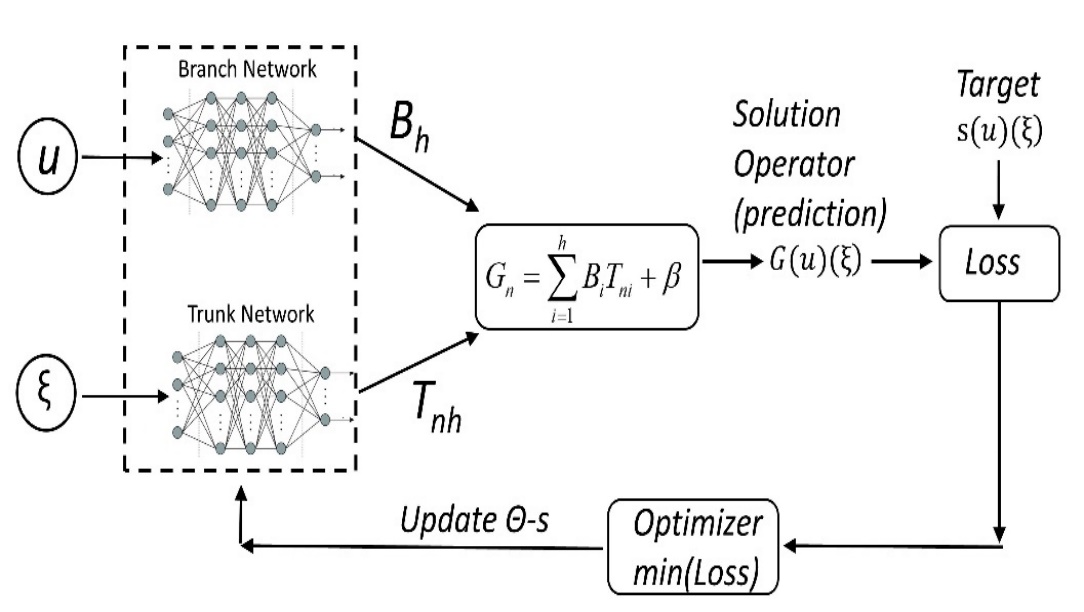}
    \caption{Original DeepONet architecture. The branch network encodes the input function to $B_h$, while the trunk network encodes spatiotemporal coordinates to $T_{nh}$. Their dot product along the hidden dimension $h$ produces the solution field. \textcolor{black}{The symbol $\theta$ represents trainable parameters contained in the model.}}
    \label{fig:3}
\end{figure}

\subsection*{Detailed model architectures}
This subsection provides full layer-by-layer specifications of the branch and trunk networks used in each benchmark, complementing the high-level description in the main manuscript. All models were implemented in DeepXDE (TensorFlow backend) with hidden dimension $h = 100$, zero-initialized biases, and a trunk MLP of widths $(101, 101, 101, 101, 101, h\cdot c)$, where $c$ denotes the number of output solution fields. Two-branch (MIONet-style) variants apply identical per-input pipelines to each input channel and merge their outputs via the Hadamard product, $B = B_d \odot B_m$. Sequential branches share a common GRU encoder–decoder backbone:
\[
\text{encoder}\,[\mathrm{GRU}^{\mathrm{seq}}_{256},\, \mathrm{GRU}_{128}]
\;\to\; \texttt{RepeatVector}(100)
\;\to\; \text{decoder}\,[\mathrm{GRU}^{\mathrm{seq}}_{128},\, \mathrm{GRU}^{\mathrm{seq}}_{256}]
\;\to\; \texttt{TD Dense}(1),
\]
where $\mathrm{GRU}^{\mathrm{seq}}_u$ returns the full sequence with $u$ units, $\mathrm{GRU}_u$ returns the final state, $\texttt{RepeatVector}(L)$ replicates the latent to length $L$, and \texttt{TD Dense} denotes a time-distributed linear head applied at each time step.

\paragraph{Reaction–diffusion (DeepONet, $c=1$).}
Branches are feedforward MLPs rather than GRUs, since the inputs $u_0(x)$ and $k(x)$ are spatial and not path-dependent. The single-branch model concatenates both inputs defined on the 255-node mesh, flattens them to 510 features, and processes the result through an MLP with widths $(512, 512, 512, 512, 256, 100)$ to produce $B \in \mathbb{R}^{100\times 1}$. The two-branch (MIONet-style) variant applies the same MLP independently to each 255-dimensional input, yielding $B_d, B_m \in \mathbb{R}^{100\times 1}$, which are merged element-wise. The trunk takes spatiotemporal coordinates $\xi=(x,t)$ and uses widths $(101, 101, 101, 101, 101, 100)$, giving $T(\xi) \in \mathbb{R}^{100\times 1}$ per query and $T \in \mathbb{R}^{n\times 100\times 1}$ over $n$ points.

\paragraph{Electro–thermal (S-DeepONet, $c=2$).}
The branch ingests the length-101, two-channel sequence $[Q_{\mathrm{ext}}(t), \rho_e(t)]$ through the encoder–decoder above, producing $B \in \mathbb{R}^{100\times 1}$. The two-branch variant applies one such pipeline per input channel (feature dimension $=1$) and merges via $B = B_d \odot B_m$. The trunk takes spatial coordinates $(x,y)$ with widths $(101, 101, 101, 101, 101, 200)$, producing $T_{hc}\in\mathbb{R}^{100\times 2}$ per query and $T_{nhc} \in \mathbb{R}^{n\times 100\times 2}$ over $n$ points.

\paragraph{Thermo–mechanical (S-DeepONet, $c=2$).}
The branch ingests the boundary histories $[q(t),\,d_n(t)]$ (length 101, two channels) through the same encoder–decoder, yielding $B \in \mathbb{R}^{100\times 1}$. The two-branch variant again applies one pipeline per input and merges via $B = B_d \odot B_m$. The trunk maps $(x,y)$ with widths $(101, 101, 101, 101, 101, 200)$, giving $h\cdot c = 200$ with $c=2$, hence $T_{hc}\in\mathbb{R}^{100\times 2}$ and $T_{nhc} \in \mathbb{R}^{n\times 100\times 2}$ over $n$ points.

\section*{Supplementary Note 2: Training and Test Log History}
\begin{figure}[!htbp]
    \centering
    \includegraphics[width=0.95\textwidth]{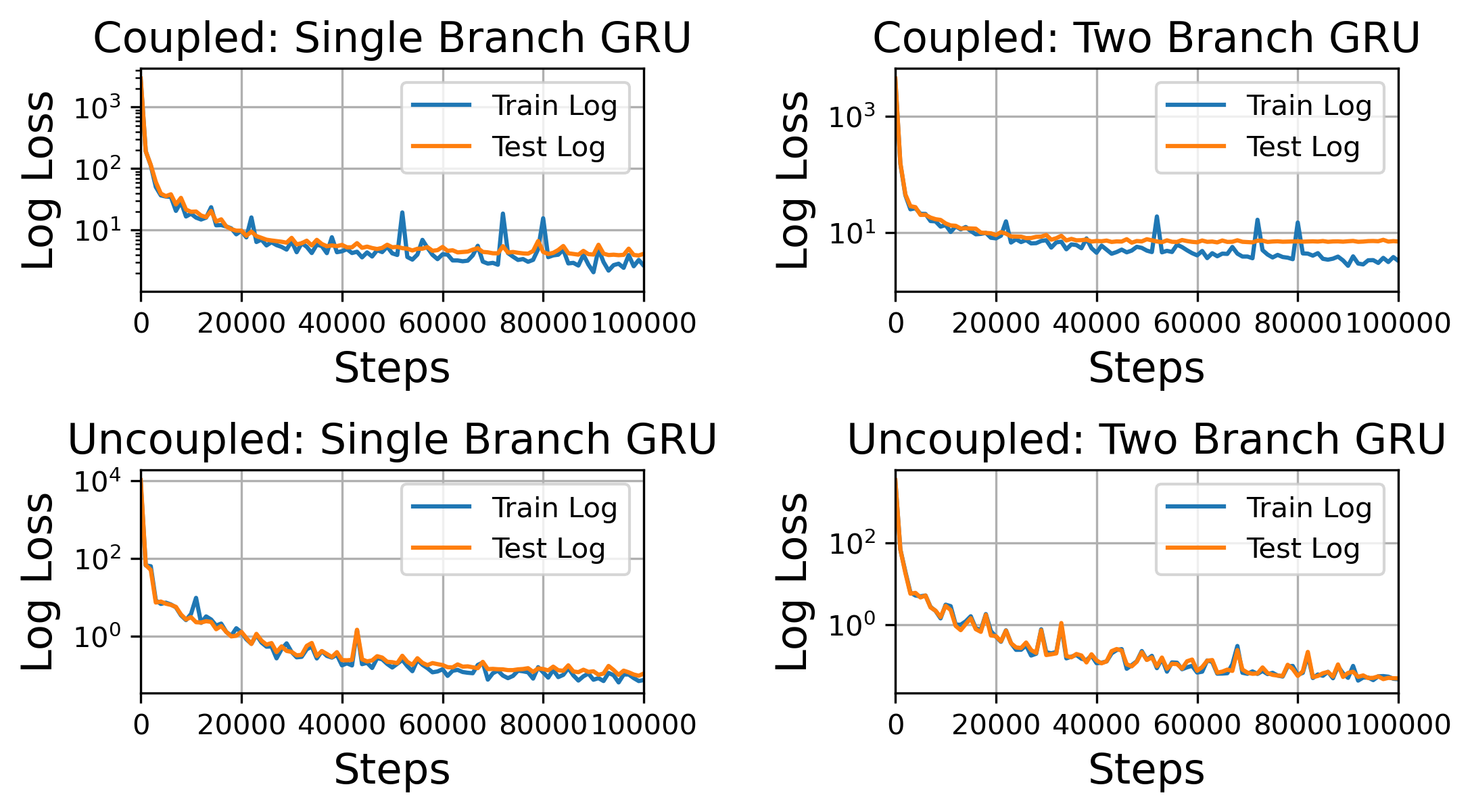}
    \caption{Training and testing log loss histories for the coupled and uncoupled electro-thermal system using GRU-based S-DeepONets. (Top Left) Coupled system with single-branch GRU, (Top Right) Coupled system with two-branch GRU, (Bottom Left) Uncoupled system with single-branch GRU, and (Bottom Right) Uncoupled system with two-branch GRU.}
    \label{fig:thermal_electric_log}
\end{figure}

\begin{figure}[!htbp]
    \centering
    \includegraphics[width=0.95\textwidth]{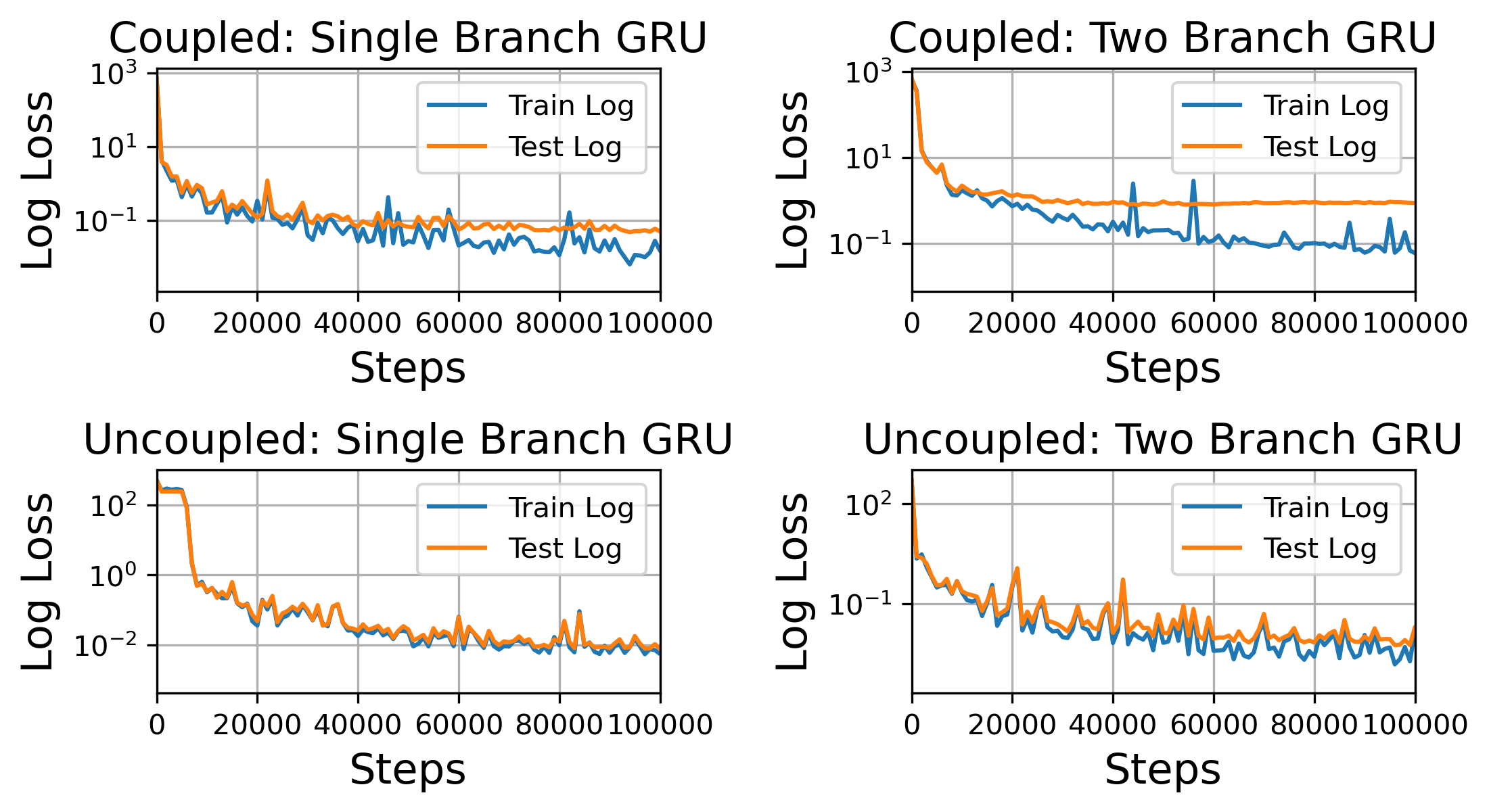}
    \caption{Training and testing log loss histories for the coupled and uncoupled thermo-mechanical system using GRU-based S-DeepONets. (Top Left) Coupled system with single-branch GRU, (Top Right) Coupled system with two-branch GRU, (Bottom Left) Uncoupled system with single-branch GRU, and (Bottom Right) Uncoupled system with two-branch GRU.}
    \label{fig:thermo_mechanical_log}
\end{figure}

To support the reproducibility and transparency of our empirical results, we provide the full training and test log loss histories for both benchmark multiphysics problems: the electro-thermal system and the thermo–mechanical system. Each experiment was conducted using GRU-based S-DeepONet architectures under both coupled and uncoupled conditions. For each configuration, we include results from both single-branch and two-branch model variants. Figures~\ref{fig:thermal_electric_log} and \ref{fig:thermo_mechanical_log} show the log-scaled training and testing loss curves over 100{,}000 optimization steps for the thermal–electric and thermo–mechanical systems, respectively. All models demonstrate stable convergence. 

\clearpage

\section*{Supplementary Note 3: Architectural Ablation}
In addition to the model architecture described in the manuscript, an ablation study was conducted in which the datasets, training protocol, loss weighting, and evaluation were held constant, and only the sequence encoder and branch design were varied. GRU and LSTM cells were compared across multiple depths and hidden dimensions. A single-branch encoder that jointly embeds all driving signals was compared with a multi-branch encoder, in which signals were embedded separately before being fused with the trunk. Performance was reported as the mean $L_2$ error per field to permit direct comparison across branch designs, sequence cells, and capacity settings.

\subsection*{Electro-thermal coupling example}
An ablation study was conducted to evaluate the impact of architectural variations on the predictive performance of the electro-thermal multiphysics system. Specifically, GRU and LSTM sequence encoders were compared across multiple network depths and hidden dimensions, under both single- and multi-branch encoder configurations. The results, summarized in Table~\ref{tab:problem2}, are reported as mean $L_2$ errors for the temperature and electric potential fields. Performance is grouped according to whether the underlying PDE system was configured in a coupled or uncoupled manner.

For the coupled system, it was consistently observed that single-branch architectures achieved lower errors than their multi-branch counterparts. The result supports the interpretation that concatenating inputs in a single branch allows the model to more effectively capture nonlinear interactions arising in coupled PDE systems. LSTM-based models exhibited similar trends but generally produced higher errors than GRUs under equivalent settings.

In contrast, for the uncoupled system, the performance differences between single- and multi-branch architectures were less consistent. While multi-branch models still tended to perform better in many cases, the gap was narrower, and in certain configurations, single-branch networks achieved comparable (or slightly better) accuracy. These phenomena was more observed in electric potential predictions.

\begin{table}[htbp]
\centering
\caption{Performance comparison of GRU and LSTM models under coupled and uncoupled settings. 
Each cell shows mean $L_2$ errors of temperature and electric potential, respectively.}
\label{tab:problem2}
\setlength{\tabcolsep}{6pt}
\resizebox{\textwidth}{!}{%
\begin{tabular}{@{}cccccc@{}}
\toprule
\multirow{2}{*}{GRU Depth} & \multirow{2}{*}{Hidden Dim (HD)} & 
\multicolumn{2}{c}{Coupled} & \multicolumn{2}{c}{Uncoupled} \\ \cmidrule(l){3-6} 
 &  & Single-Branch & Multi-Branch & Single-Branch & Multi-Branch \\ 
 &  &  (\% / \%)    &  (\% / \%)   &  (\% / \%)   & (\% / \%) \\
 \midrule \midrule
2 & 100 & \makecell{$0.56$ / $1.16$} & \makecell{$3.57$ / $5.58$} & \makecell{$0.07$ / $0.96$} & \makecell{$0.03$ / $0.59$} \\
4 & 50  & \makecell{$0.72$ / $0.94$} & \makecell{$5.53$ / $16.54$} & \makecell{$0.05$ / $4.84$} & \makecell{$0.07$ / $2.22$} \\
(4) & (100) & \makecell{$0.50$ / $1.10$} & \makecell{$6.61$ / $15.50$} & \makecell{$0.13$ / $0.97$} & \makecell{$0.07$ / $0.80$} \\
4 & 150 & \makecell{$0.59$ / $0.95$} & \makecell{$6.47$ / $21.20$} & \makecell{$0.09$ / $0.86$} & \makecell{$0.04$ / $0.58$} \\
4 & 200 & \makecell{$0.49$ / $0.87$} & \makecell{$4.22$ / $7.57$} & \makecell{$0.09$ / $0.59$} & \makecell{$0.08$ / $0.49$} \\
6 & 100 & \makecell{$0.58$ / $0.97$} & \makecell{$5.90$ / $13.52$} & \makecell{$0.06$ / $0.44$} & \makecell{$0.05$ / $0.97$} \\
8 & 100 & \makecell{$0.56$ / $0.94$} & \makecell{$5.77$ / $15.39$} & \makecell{$0.05$ / $0.50$} & \makecell{$0.11$ / $1.21$} \\ \midrule
\multicolumn{6}{l}{LSTM Depth} \\ 
4 & 100 & \makecell{$0.82$ / $1.25$} & \makecell{$5.45$ / $8.08$} & \makecell{$0.085$ / $1.28$} & \makecell{$0.046$ / $1.82$} \\ 
\bottomrule
\end{tabular}%
}
\end{table}

\subsection*{Thermo-mechanical solidification example}
The same study was conducted for the thermo-mechanical solidification problem under a different type of multiphysics coupling. As with the electro-thermal case, GRU and LSTM encoders were tested across various network depths and hidden dimensions, using both single- and multi-branch encoder configurations. The same dataset, training procedure, and evaluation protocol were applied uniformly to ensure comparability. Table~\ref{tab:problem3} presents the mean $L_2$ errors for temperature and stress predictions under coupled and uncoupled conditions. Results are grouped by model depth and hidden dimension, with comparisons between GRU- and LSTM-based networks.

\begin{table}[htbp]
\centering
\caption{Performance comparison of GRU and LSTM models under coupled and uncoupled settings. 
Each cell shows the MAE of [temperature/stress], respectively. Units in the table are [${ }^{\circ} \mathrm{C}$ / MPa].}
\label{tab:problem3}
\setlength{\tabcolsep}{6pt}
\resizebox{\textwidth}{!}{%
\begin{tabular}{@{}cccccc@{}}
\toprule
\multirow{2}{*}{GRU Depth} & \multirow{2}{*}{Hidden Dim (HD)} & 
\multicolumn{2}{c}{Coupled} & \multicolumn{2}{c}{Uncoupled} \\ \cmidrule(l){3-6} 
 &  & Single-Branch & Multi-Branch & Single-Branch & Multi-Branch \\ 
 &  &  (${ }^{\circ} \mathrm{C}$ / MPa)    &  (${ }^{\circ} \mathrm{C}$ / MPa)   &  (${ }^{\circ} \mathrm{C}$ / MPa)   & (${ }^{\circ} \mathrm{C}$ / MPa) \\
  \midrule \midrule
2 & 100 & \makecell{1.20 / 0.13} & \makecell{2.03 / 0.32} & \makecell{0.69 / 0.0097} & \makecell{0.68 / 0.0070} \\
4 & 50  & \makecell{1.02 / 0.13} & \makecell{2.02 / 0.32} & \makecell{0.89 / 0.0074} & \makecell{0.58 / 0.0085} \\
(4) & (100) & \makecell{0.91 / 0.12} & \makecell{1.56 / 0.23} & \makecell{0.77 / 0.0089} & \makecell{0.75 / 0.0053} \\
4 & 150 & \makecell{0.84 / 0.12} & \makecell{2.56 / 0.32} & \makecell{0.76 / 0.0094} & \makecell{0.65 / 0.0063} \\
4 & 200 & \makecell{0.89 / 0.10} & \makecell{1.24 / 0.24} & \makecell{1.49 / 0.0110} & \makecell{1.02 / 0.0330} \\
6 & 100 & \makecell{1.14 / 0.11} & \makecell{1.39 / 0.25} & \makecell{1.01 / 0.0059} & \makecell{0.98 / 0.0051} \\
8 & 100 & \makecell{0.99 / 0.11} & \makecell{1.96 / 0.25} & \makecell{0.76 / 0.0065} & \makecell{0.52 / 0.0067} \\ \midrule
\multicolumn{6}{l}{LSTM Depth} \\ 
4 & 100 & \makecell{0.67 / 0.13} & \makecell{0.67 / 0.21} & \makecell{0.72 / 0.0067} & \makecell{0.57 / 0.0052} \\ 
\bottomrule
\end{tabular}%
}
\end{table}

For the coupled setting, single-branch GRU models consistently demonstrated lower prediction errors than their multi-branch counterparts, particularly for the stress field. Multi-branch networks exhibited large error variability and pronounced degradation in several cases, likely due to challenges in capturing tightly coupled thermo-mechanical interactions when inputs are processed independently. LSTM-based models showed comparable trends, although their stress prediction errors were generally higher overall.

In the uncoupled setting, performance differences between branch designs were less stark. While the multi-branch GRU models maintained favorable performance, single-branch networks were competitive (and slightly better) in some configurations, particularly for stress predictions. This outcome is consistent with the hypothesis that decoupled systems impose less representational burden on the encoder, thereby reducing the penalty of input segregation.

\newpage


\section*{Supplementary Note 4: Sensitivity to Training Set Size}
To estimate the amount of data needed, a study was conducted in which only the training set size (20\%, 50\%, 70\%, and 100\%) was varied, while the model, training setup, and evaluation remained constant. For each size, subsets were drawn without replacement, and test performance was summarized as mean $L_2$ error per field under the same coupled/uncoupled and single-/multi-branch settings; the same steps were applied to both the electro-thermal (temperature, electric potential) and thermo-mechanical solidification (temperature, stress) examples.

\begin{table}[htbp]
\centering
\caption{Effect of training datasets on the model performance in the electro-thermal coupling example. Each cell shows the mean relative $L_2$ errors of [temperature / electric potential]. Units in the table are \%. }
\begin{tabular}{@{}ccccc@{}}
\toprule
\multirow{2}{*}{Training Dataset Size (\%)} & \multicolumn{2}{c}{Coupled} & \multicolumn{2}{c}{Uncoupled} \\ \cmidrule(l){2-5} 
 & Single-Branch & Multi-Branch & Single-Branch & Multi-Branch \\ 
 &  (\% / \%)    &  (\% / \%)   &  (\% / \%)   & (\% / \%) \\
 \midrule \midrule
20 & \makecell{$2.07$ / $4.25$} & \makecell{$8.29$ / $20.9$} & \makecell{$0.15$ / $1.10$} & \makecell{$0.05$ / $4.74$} \\
50 & \makecell{$0.70$ / $1.92$} & \makecell{$3.51$ / $6.51$} & \makecell{$0.08$ / $0.51$} & \makecell{$0.08$ / $4.75$} \\
70 & \makecell{$0.62$ / $1.38$} & \makecell{$4.52$ / $7.87$} & \makecell{$0.05$ / $0.32$} & \makecell{$0.14$ / $0.33$} \\
(100) & \makecell{$0.50$ / $1.10$} & \makecell{$6.61$ / $15.50$} & \makecell{$0.13$ / $0.97$} & \makecell{$0.07$ / $0.80$} \\ \bottomrule
\end{tabular}
\end{table}

\begin{table}[htbp]
\centering
\caption{Effect of training datasets on the model performance in thermo-mechanical solidification example. Each cell shows the MAE of [temperature/stress], respectively. Units in the table are [${ }^{\circ} \mathrm{C}$ / MPa]. }
\begin{tabular}{@{}ccccc@{}}
\toprule
\multirow{2}{*}{Training Dataset Size (\%)} & \multicolumn{2}{c}{Coupled} & \multicolumn{2}{c}{Uncoupled} \\ \cmidrule(l){2-5} 
 & Single-Branch & Multi-Branch & Single-Branch & Multi-Branch \\ 
 &  (${ }^{\circ} \mathrm{C}$ / MPa)    &  (${ }^{\circ} \mathrm{C}$ / MPa)   &  (${ }^{\circ} \mathrm{C}$ / MPa)   & (${ }^{\circ} \mathrm{C}$ / MPa) \\
 \midrule \midrule
20 & \makecell{3.93 / 0.29} & \makecell{1.16 / 0.47} & \makecell{1.30 / 0.0130} & \makecell{1.01 / 0.0092} \\
50 & \makecell{0.95 / 0.17} & \makecell{0.86 / 0.26} & \makecell{0.63 / 0.0078} & \makecell{0.53 / 0.0050}  \\
70 & \makecell{0.87 / 0.15} & \makecell{0.87 / 0.27} & \makecell{0.65 / 0.0080} & \makecell{0.60 / 0.0054} \\
(100) & \makecell{0.91 / 0.12} & \makecell{1.56 / 0.23} & \makecell{0.77 / 0.0089} & \makecell{0.75 / 0.0053} \\ \bottomrule
\end{tabular}
\end{table}

\newpage

\section*{Supplementary Note 5: Robustness Across Independent Runs (Mean $\pm$ Std Over 5 Seeds)}
To assess reproducibility, we repeated training and evaluation five times with different random seeds while keeping hyperparameters, and training schedules fixed. We report the \textbf{mean $\pm$ standard deviation} of errors for each case across the five runs; lower is better. Boldface marks the better architecture within each row. 
Here, ``Single-branch'' denotes a shared-branch design, and ``Two-branch (MIONet-style)'' denotes a dual-branch design. ``Coupled'' indicates physically coupled fields solved jointly; ``Uncoupled'' indicates the fields are generated in isolation.

\begin{table}[htbp]
\caption{Mean $L_2$ relative errors for \textbf{Temperature} (\%, mean $\pm$ std over 5 runs).}
\centering
\begin{tabular}{@{}lcc@{}}
\toprule
$L_2$ relative error {[}\%{]} & Single-branch & Two-branch (MIONet-style) \\ \midrule
Coupled physics   & \textbf{0.526 $\pm$ 0.038} & 3.807 $\pm$ 0.856 \\
Uncoupled physics & 0.081 $\pm$ 0.0313          & \textbf{0.070 $\pm$ 0.021} \\ \bottomrule
\end{tabular}
\label{tab:supp_thermoelectric_temperature}
\end{table}

\begin{table}[htbp]
\caption{Mean $L_2$ relative errors for \textbf{Electrical Potential} (\%, mean $\pm$ std over 5 runs).}
\centering
\begin{tabular}{@{}lcc@{}}
\toprule
$L_2$ relative error {[}\%{]} & Single-branch & Two-branch (MIONet-style) \\ \midrule
Coupled physics   & \textbf{0.763 $\pm$ 0.176} & 4.227 $\pm$ 1.002 \\
Uncoupled physics & 0.551 $\pm$ 0.325           & \textbf{0.370 $\pm$ 0.279} \\ \bottomrule
\end{tabular}
\label{tab:supp_thermoelectric_potential}
\end{table}

\begin{table}[htbp]
\caption{\textbf{Stress} MAE (MPa, mean $\pm$ std over 5 runs).}
\centering
\begin{tabular}{@{}lcc@{}}
\toprule
MAE {[}MPa{]} & Single-branch & Two-branch (MIONet-style) \\ \midrule
Coupled physics   & \textbf{0.120 $\pm$ 0.007} & 0.219 $\pm$ 0.022 \\
Uncoupled physics & 0.008 $\pm$ 0.001          & \textbf{0.006 $\pm$ 0.001} \\ \bottomrule
\end{tabular}
\label{tab:supp_thermomech_stress}
\end{table}

\begin{table}[H]
\caption{\textbf{Temperature} MAE ($^{\circ}$C, mean $\pm$ std over 5 runs).}
\centering
\begin{tabular}{@{}lcc@{}}
\toprule
MAE {[}$^{\circ}$C{]} & Single-branch & Two-branch (MIONet-style) \\ \midrule
Coupled physics   & \textbf{0.869 $\pm$ 0.097} & 1.466 $\pm$ 0.439 \\
Uncoupled physics & 0.703 $\pm$ 0.176 & \textbf{0.572 $\pm$ 0.185} \\ \bottomrule
\end{tabular}
\label{tab:supp_thermomech_temperature}
\end{table}

Across five independent runs, the variability-aware summary is consistent with the main manuscript: under coupled physics, the single-branch model achieves lower errors, whereas under uncoupled physics, the two-branch (MIONet-style) tends to perform better. The magnitude of the effect differs by regime: in coupled settings, the single-branch advantage is clear and stable across seeds, while in uncoupled settings, the two-branch advantage is present but comparatively modest. Detailed figures are reported in Tables~\ref{tab:supp_thermoelectric_temperature}–\ref{tab:supp_thermomech_temperature}.

\newpage

\section*{Supplementary Note 6: Robustness of the model against Noisy Inputs}
\textcolor{black}{
In practical monitoring and control settings, the input quantities driving the system are not directly observed as exact functions, but are instead inferred from sensor measurements that are subject to noise and uncertainty. Assessing the robustness of a learned operator to such perturbations, and quantifying the resulting uncertainty in its predictions, is therefore essential for evaluating its suitability for deployment. To address this, we evaluate robustness and uncertainty propagation using the coupled, single-branch S-DeepONet model introduced for the thermo-mechanical problem given in the main paper, section 2.3 and 5.1.}

\begin{figure}[!htbp]
    \centering
    \begin{subfigure}[b]{0.99\textwidth}
        \centering
        \includegraphics[width=\textwidth]{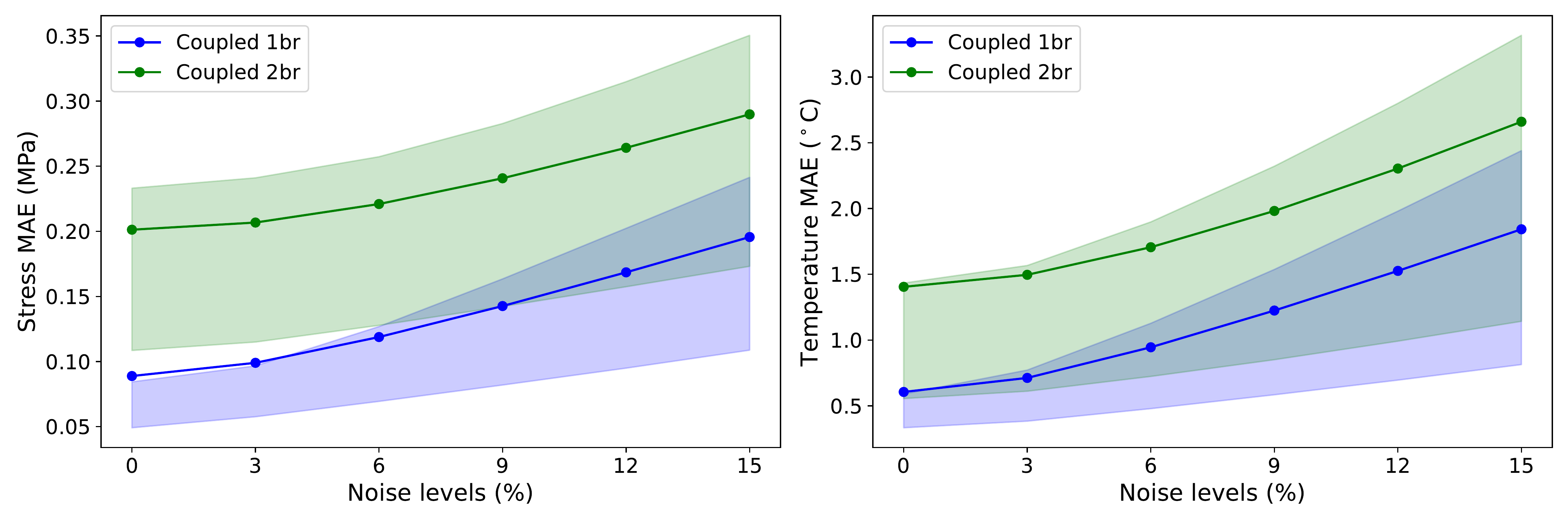}
        \caption{\textcolor{black}{Couple physics - MAE on stress and temperature on different noise levels added to the input}}
        \label{fig:img1}
    \end{subfigure}
    \vspace{1em} 
    
    \begin{subfigure}[b]{0.99\textwidth}
        \centering
        \includegraphics[width=\textwidth]{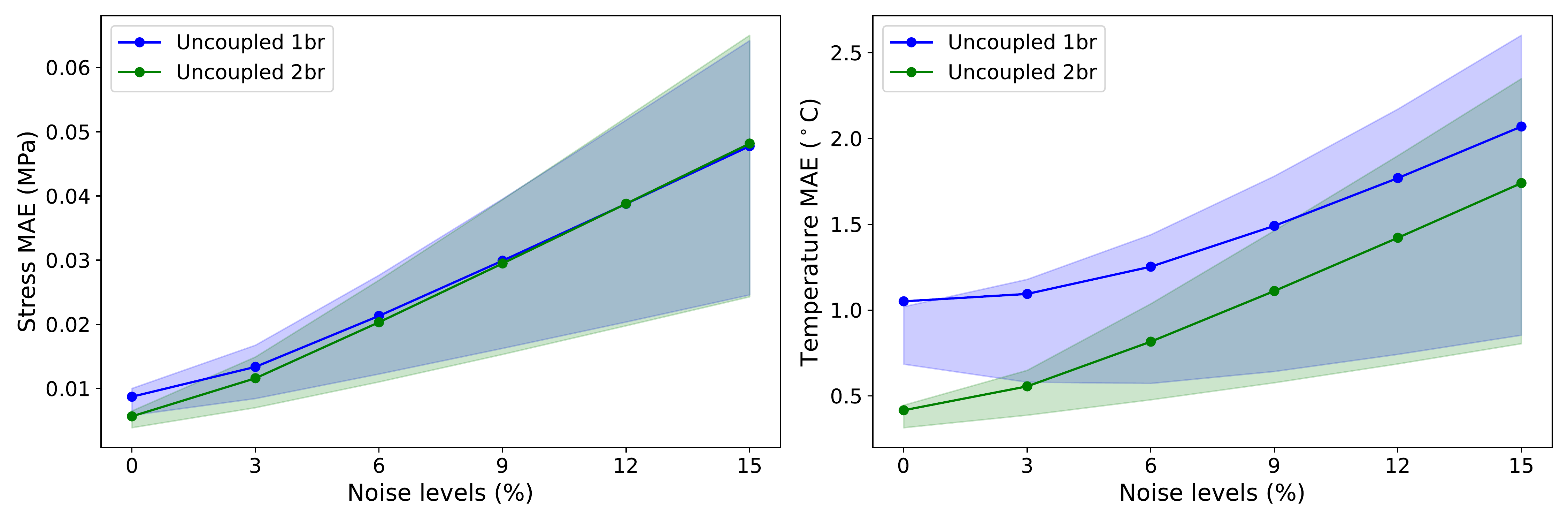}
        \caption{\textcolor{black}{Uncouple physics - MAE on stress and temperature on different noise levels added to the input}}
        \label{fig:img3}
    \end{subfigure}
    
    \caption{\textcolor{black}{Mean absolute error (MAE) of stress and temperature predictions under coupled / uncoupled physics on thermo-mechanical model varying input noise levels for the 1-branch (1br) and 2-branch (2br) architectures. Dots are the mean value of each model at the noise level. Shaded regions indicate the inter-quartile range (IQR) across test samples.}}

    \label{fig:overall_comparison}
\end{figure}

\textcolor{black}{Table ~\ref{tab:stress_noise} and Table ~\ref{tab:temperature_noise} summarizes the robustness of the thermo–mechanical neural operator under increasing levels of measurement noise injected into the input functions. For each noise level (3-15\%), $M=100$ noisy realizations of each test input function were generated and propagated through the pre-trained model, and MAE was aggregated across test samples. Figure~\ref{fig:overall_comparison} visualizes the effect of noisy input on MAE, the shaded regions indicate the interquartile range (IQR), reflecting variability across physical test cases.}

\begin{table}[h]
\centering
\caption{Stress values across different noise levels}
\label{tab:stress_noise}
\begin{tabular}{lcccccc}
\toprule
Configuration & 0.0\% & 3\% & 6\% & 9\% & 12\% & 15\% \\
\midrule
1 branch coupled   & 0.0889 & 0.0990 & 0.1188 & 0.1426 & 0.1685 & 0.1956 \\
2 branch coupled   & 0.2012 & 0.2067 & 0.2210 & 0.2407 & 0.2641 & 0.2898 \\
1 branch uncoupled & 0.0087 & 0.0134 & 0.0213 & 0.0299 & 0.0388 & 0.0478 \\
2 branch uncoupled & 0.0057 & 0.0116 & 0.0203 & 0.0295 & 0.0388 & 0.0482 \\
\bottomrule
\end{tabular}
\end{table}

\begin{table}[h]
\centering
\caption{Temperature values across different noise levels}
\label{tab:temperature_noise}
\begin{tabular}{lcccccc}
\toprule
Configuration & 0.0\% & 3\% & 6\% & 9\% & 12\% & 15\% \\
\midrule
1 branch coupled   & 0.6059 & 0.7126 & 0.9455 & 1.2247 & 1.5262 & 1.8425 \\
2 branch coupled   & 1.4050 & 1.4964 & 1.7061 & 1.9818 & 2.3042 & 2.6600 \\
1 branch uncoupled & 1.0512 & 1.0943 & 1.2531 & 1.4910 & 1.7694 & 2.0698 \\
2 branch uncoupled & 0.4164 & 0.5559 & 0.8162 & 1.1114 & 1.4216 & 1.7402 \\
\bottomrule
\end{tabular}
\end{table}

\textcolor{black}{Under coupled physics, both the 1-branch and 2-branch architectures exhibit an increase in MAE for stress and temperature as the noise level increases. It indicates graceful degradation under input perturbations. Across all noise levels, the 1-branch model consistently achieves lower MAE and narrower IQRs than the 2-branch model, suggesting improved robustness when thermo–mechanical interactions are encoded through a shared latent representation.}

\textcolor{black}{In the uncoupled setting, a more nuanced behavior is observed. At low noise levels, the 2-branch model outperforms the 1-branch model in stress prediction, consistent with earlier baseline results and reflecting the increased representational capacity afforded by separate encoders. However, as the noise level increases, the performance gap narrows, and beyond approximately $12\%$ noise, the 1-branch model becomes more robust and ultimately surpasses the 2-branch model. This crossover behavior suggests that, while the 2-branch architecture is advantageous in low-noise scenarios, it is more sensitive to measurement noise in the uncoupled setting, whereas the shared-branch formulation provides an implicit regularization effect that improves robustness under degraded input quality.}

\textcolor{black}{Taken together, these results highlight a tradeoff between expressiveness and robustness in multi-branch operator architectures. While separate branches can better exploit clean, modality-specific information, shared representations appear more resilient to noisy or uncertain inputs, particularly when physical coupling is weak or absent.}

\end{document}